\pdfoutput=1
\ifdefined\pdfminorversion\pdfminorversion=7\fi
\ifdefined\pdfsuppressptexinfo\pdfsuppressptexinfo=-1\fi
\documentclass{article}
\PassOptionsToPackage{table}{xcolor}
\usepackage{iclr2027_conference,times}

\usepackage{amsmath,amsfonts,bm}

\def\1{\bm{1}}

\DeclareMathAlphabet{\mathsfit}{\encodingdefault}{\sfdefault}{m}{sl}
\SetMathAlphabet{\mathsfit}{bold}{\encodingdefault}{\sfdefault}{bx}{n}

\usepackage{hyperref}
\usepackage{url}
\usepackage{xurl}
\usepackage{fontawesome5}
\usepackage{booktabs}
\usepackage{tabularx}
\usepackage{placeins}
\usepackage{graphicx}
\usepackage{subcaption}
\usepackage{wrapfig}
\usepackage{needspace}
\usepackage{multirow}

\makeatletter
\renewcommand{\fps@figure}{t}
\makeatother

\usepackage[table]{xcolor}

\definecolor{paperlink}{HTML}{0071BC}
\definecolor{paperref}{HTML}{ED1C24}
\hypersetup{colorlinks=true,citecolor=paperlink,linkcolor=paperref,urlcolor=paperlink,pdftitle={AdaLoop: Thinking with the Vision Encoder},pdfauthor={Xinran Nicole Han, Yiyang Lu, Katelyn Gan, Todd Zickler}}

\title{AdaLoop: Thinking with the Vision Encoder}

\author{
Xinran Nicole Han$^{1}$\thanks{Equal contribution.}%
\hspace{0.5em}%
\thanks{Corresponding author: \texttt{xinranhan@g.harvard.edu}.}
\quad
Yiyang Lu$^{2}$\footnotemark[1]
\quad
Katelyn Gan$^{2}$ \quad
Todd Zickler$^{1}$ \\
\textnormal{$^{1}$Harvard University \qquad $^{2}$MIT}
}

\iclrfinalcopy
\begin{document}

\maketitle
\lhead{Preprint}

\newskip\arxivSavedparskip
\arxivSavedparskip=\parskip
\setlength{\parskip}{5.5pt plus 1pt minus 1pt}
\newskip\arxivSavedtextfloatsep
\arxivSavedtextfloatsep=\textfloatsep
\setlength{\textfloatsep}{14pt plus 2pt minus 2pt}
\newskip\arxivSavedfloatsep
\arxivSavedfloatsep=\floatsep
\setlength{\floatsep}{10pt plus 2pt minus 2pt}
\newskip\arxivSavedabovedisplayskip
\arxivSavedabovedisplayskip=\abovedisplayskip
\setlength{\abovedisplayskip}{7pt plus 2pt minus 1pt}
\newskip\arxivSavedbelowdisplayskip
\arxivSavedbelowdisplayskip=\belowdisplayskip
\setlength{\belowdisplayskip}{7pt plus 2pt minus 1pt}
\newskip\arxivSavedabovedisplayshortskip
\arxivSavedabovedisplayshortskip=\abovedisplayshortskip
\setlength{\abovedisplayshortskip}{3pt plus 1pt}
\newskip\arxivSavedbelowdisplayshortskip
\arxivSavedbelowdisplayshortskip=\belowdisplayshortskip
\setlength{\belowdisplayshortskip}{5pt plus 1pt minus 1pt}
\captionsetup[figure]{skip=8pt}
\let\arxivSavedNormalsize\normalsize
\renewcommand{\normalsize}{%
  \arxivSavedNormalsize
  \setlength{\abovedisplayskip}{7pt plus 2pt minus 1pt}%
  \setlength{\belowdisplayskip}{7pt plus 2pt minus 1pt}%
  \setlength{\abovedisplayshortskip}{3pt plus 1pt}%
  \setlength{\belowdisplayshortskip}{5pt plus 1pt minus 1pt}%
}
\normalsize

\begin{abstract}
Most vision-language models (VLMs) treat the vision encoder as a static feature extractor: it encodes an image once, independently of the question, and leaves subsequent reasoning to the language model. Yet different questions may require different visual operations and processing depths. We present AdaLoop, a framework for question-conditioned visual recurrence in VLMs. AdaLoop adapts visual computation to the question and reuses the encoder's transformations across multiple iterations, with the pretrained vision weights frozen. We also introduce VisRO, a procedural dataset and generation pipeline for training and evaluating vision-centric reasoning. On six VisRO tasks, AdaLoop outperforms standard fine-tuning and question-conditioned single-pass baselines with matched training data and optimizer updates. These gains hold across three VLM families on both in-distribution and out-of-distribution examples. Our findings suggest that pretrained vision encoders can contribute to reasoning through task-directed, iterative visual computation.

\end{abstract}

\vspace{-0.75em}
\begingroup
\centering\small
\renewcommand{\arraystretch}{1.1}
\begin{tabular}{@{}c@{\hspace{0.6em}}l@{\hspace{2em}}l@{}}
\faGithub & \textbf{Code}  & \url{https://github.com/xrhan/AdaLoop} \\
\faGithub & \textbf{VisRO} & \url{https://github.com/Lyy-iiis/VisRO} \\
\end{tabular}\par
\endgroup
\vspace{0.4em}

\section{Introduction}
\label{sec:intro}

Most vision--language models (VLMs) encode an image once, independently of the question, and pass the resulting visual tokens to a language model for subsequent reasoning. The same visual computation serves every question, whether the answer depends on a single local feature or on relating distant parts of the scene. A fixed pass may not perform the computation a task requires: useful visual features may lie at intermediate encoder layers~\citep{bolya2026perception} or be lost along the way~\citep{tong2024eyes,ganz2024qavit}. The encoder thus acts only as a feature extractor, even though vision architectures can themselves learn to reason when trained from scratch~\citep{hu2026arc,shu2026loopvit}. Visual reasoning, then, need not be left entirely to the language model.

For a pretrained VLM, an appealing possibility is to reuse the visual transformations already learned by its vision encoder. We draw inspiration from human vision, where processing is directed by the task~\citep{ullman1984} and recurrent~\citep{lamme2000distinct,rao1999predictive}. We investigate how to equip a pretrained vision encoder with both capabilities through (i) \emph{question conditioning}, which modulates the encoder's features according to the task~\citep{perez2018film}, and (ii) \emph{looping (or recurrence)}, which reapplies weight-tied transformations to an evolving state~\citep{universal_transformers, geiping2026scaling,zhu2025ouro}.

Introducing these mechanisms into a pretrained VLM raises two questions. First, each encoder layer was trained to operate at a particular feedforward depth, so simply looping a span of its frozen layers yields little benefit (Section~\ref{sec:recurrence-ablation}). Second, looping produces a sequence of visual states rather than a single output, and it is not obvious how to present the sequence to the language model.

\begin{figure}[t]
  \centering
  \includegraphics[width=0.98\linewidth]{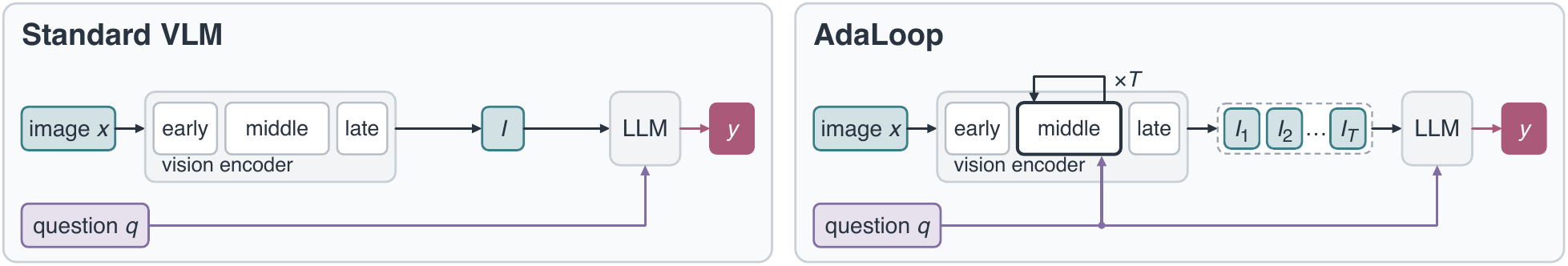}
  \caption{\textbf{Looking once versus repeatedly.}
  \emph{Left:} A standard VLM encodes the image once, independently
  of the question.
  \emph{Right:} AdaLoop turns a span of the vision encoder into a
  question-conditioned \emph{perception loop}, updating its visual
  state over $T$ iterations. The language model answers from the
  resulting sequence, using either the current state or retained
  history.}
  \label{fig:teaser}
\end{figure}

We approach this as a design problem and present AdaLoop, a framework for adaptive visual recurrence in pretrained VLMs (Figure~\ref{fig:teaser}). AdaLoop creates a \emph{perception loop} by repeatedly applying a span of the vision encoder to an evolving visual state, while keeping the original vision weights frozen. It uses question-conditioned adaptive layer normalization (AdaLN)~\citep{perez2018film,peebles2023scalable} to modulate the shared transformations at each iteration. We compare three ways for the language model to use the resulting sequence of states: reading only the current state, accumulating states across iterations, and interleaving states with answer drafts. All variants train only on triples of (image, question, answer), using the same final-answer target across visual iterations to support adaptive inference-time compute.

Evaluating vision-centric reasoning requires diagnostic tasks with systematic difficulty control and scalable training data. We introduce VisRO (Visual Routines) to support this research effort. VisRO is a procedural dataset and generation pipeline encompassing tasks like grouping, counting, connectivity, spatial tracing, and amodal perception---all of which demand careful visual inspection with minimal reliance on language priors or external knowledge (Figure~\ref{fig:benchmark}). Our pipeline provides programmatically verifiable answers, controllable difficulty, and controlled distributional shifts in task structure or visual appearance for out-of-distribution evaluation. VisRO serves as reusable infrastructure for the community, and our AdaLoop experiments use six tasks from the collection.

Experiments with three VLM families, Qwen2.5-VL~\citep{bai2025qwen25vl}, PaliGemma~2~\citep{steiner2024paligemma2} and LLaVA-1.5~\citep{liu2024improved} show that with matched training data and optimizer updates, AdaLoop's components provide complementary gains. Question-conditioning establishes strong single-pass controls, and looping adds a further $2.8$--$4.6$ percentage points of in-distribution VisRO accuracy, for a total
improvement of about $11$--$15$ points over standard fine-tuning. On Qwen2.5-VL, at matched parameter count and approximately matched training FLOPs, AdaLoop achieves competitive VisRO performance with better general visual question answering (VQA) capabilities than longer-trained single-pass controls. Our results suggest that existing visual transformations can be steered by the question and reused for iterative reasoning, even with the original weights frozen.

\section{Related Work}
\label{sec:related}

\paragraph{Task-directed vision.}
Task-directed processing has roots in \emph{visual routines}, combinations of basic
operations selected by the task~\citep{ullman1984}. Language can supply this
task signal through feature-wise modulation~\citep{perez2018film} or
instruction-aware feature extraction~\citep{dai2023instructblip}, and prior
work conditions vision encoders on language through
normalization~\citep{devries2017modulating,liu2026igvlm},
cross-attention~\citep{visfocus2024}, or token
injection~\citep{ganz2024qavit,tabflash2025,live2026}. AdaLoop applies such
conditioning to encoder layers that are reused across iterations.

\paragraph{Recurrent and looped architectures.}
Recurrence has a long history in computational models~\citep{pinheiro2014recurrent,liang2015recurrent,liao2016bridging,linsley2018learning}. Looped Transformers increase effective depth by sharing parameters across iterations~\citep{universal_transformers,giannou2023looped,lee2026soft}. Most are trained from scratch, either as task-specific reasoners for problems such as Sudoku, mazes, and ARC~\citep{wang2025hrm,jolicoeur2025trm,baek2026generative,shu2026loopvit} or as language models~\citep{geiping2026scaling,zhu2025ouro}. Related work studies their scaling~\citep{alabdulmohsin2025recursive,lee2026sparse,wang2026smelt,prairie2026parcae,schwethelm2026much,chen2026model}, architectural design choices~\citep{huang2026loopedmodels}, adaptive stopping~\citep{graves2016adaptive,zhou2020bert,bae2025mixture}, and iteration-dependent modulation~\citep{jeddi2026loopformer}. Other work retrofits recurrence into pretrained language models~\citep{bae2024relaxed,mcleish2025retrofitted,park2026loopus}, and recent multimodal systems loop all or part of their language backbones~\citep{xu2026looping,hive2026,park2026cvrr}. AdaLoop instead loops layers of the vision encoder in a pretrained VLM, and studies how the language model should use the resulting visual states (Appendix~\ref{app:looped-comparison}).

\paragraph{Visual reasoning in VLMs.}
Visual reasoning methods differ in where intermediate visual computation takes place and how it is controlled. External image operations and visual prompting provide selected or annotated visual evidence for language-model reasoning~\citep{shao2024visual,yang2023set,hu2024visual,wu2024v,shen2025zoomeye,zheng2026deepeyes}.
Other approaches perform intermediate computation within the language model using generated visual latents~\citep{li2025lvr,lantern2026,wang2026monet} or learnable visual tokens~\citep{li2026latent}.
Language model states or generated sub-questions can also guide further visual processing~\citep{yu2025vpt,galatolo2026lightweight,zhang2026dvr}.
AdaLoop carries an internal vision-encoder state across iterations, updating it under the guidance of the original question.

\paragraph{Visual reasoning benchmarks.}
Existing benchmarks assess broad multimodal capabilities~\citep{liu2024mmbench,chen2024we}, compositional reasoning~\citep{johnson2017clevr,hudson2019gqa}, and reasoning about text in images~\citep{singh2019towards}. BLINK and CV-Bench probe core perceptual abilities~\citep{fu2024blink,tong2024cambrian}, while BabyVision emphasizes visual reasoning with limited reliance on language knowledge~\citep{chen2026babyvision}. Scalable training resources for diverse vision-centric reasoning tasks remain limited. VisRO complements existing benchmarks with a procedural generation pipeline supporting both training and evaluation, with explicit controls over difficulty and distribution shifts.

\section{Method}

\label{sec:method}

AdaLoop introduces question-conditioned recurrence into a pretrained vision--language model. We denote its vision encoder, connector, and language model by $E$, $M$, and $G$, respectively. Given an image $x$ and a question $q$, AdaLoop repeatedly applies a group of consecutive layers in $E$ to update a visual state, with each update conditioned on $q$. The later encoder layers and the connector $M$ convert each updated state into visual tokens for $G$. We study two design choices: how to condition the visual updates on the question, and how the language model should use the visual tokens produced across iterations.

\subsection{Looped Vision Encoder}

\label{sec:vision-loop}

We organize the visual computation into three parts: a prelude $\mathcal{P}$, a loop body $\mathcal{R}$, and a coda $\mathcal{C}$. The prelude processes the image once to produce an initial visual state $z_0$. The loop body consists of consecutive encoder layers, augmented with question conditioning and reused across iterations.
Let $z_t$ denote the visual state after iteration $t$. The coda comprises the remaining encoder layers and the vision--language connector, mapping $z_t$ to a sequence of visual tokens $I_t$ for the language model,
\begin{gather}
  z_0 = \mathcal{P}(x), \qquad
  z_t = \mathcal{R}(z_{t-1};\,q), \qquad
  I_t = \mathcal{C}(z_t), \qquad t=1,\ldots,T.
  \label{eq:perception-loop}
\end{gather}
The loop body shares its parameters across iterations. The state $z_t$ carries information to the next visual iteration, while $I_t$ provides the corresponding visual input to the language model.

We condition the loop body $\mathcal{R}$ on the question $q$ through adaptive layer normalization (AdaLN)~\citep{perez2018film,peebles2023scalable}. A learned attention pooler summarizes the question's input embeddings into a vector $c_q$. For each block $\ell$, a module $A_\ell(c_q)$ predicts feature-wise shifts $\beta_\ell^f$, scales $\gamma_\ell^f$, and residual gains $\alpha_\ell^f$ for each sublayer $f\in\{\mathrm{Attn},\mathrm{MLP}\}$:
\begin{equation}
  h \leftarrow h + (\mathbf{1}+\alpha_\ell^f)\odot
  f\!\left((\mathbf{1}+\gamma_\ell^f)\odot
  \operatorname{LN}(h)+\beta_\ell^f\right),
  \label{eq:adaln}
\end{equation}
Here $h$ denotes the visual state, $\operatorname{LN}$ layer normalization, and $\odot$ element-wise multiplication. Modulation is shared across tokens and iterations. Initializing $\alpha=\beta=\gamma=0$ preserves the original encoder computation at the first training iteration for loop at $t=1$.

\paragraph{Adaptation and recurrence are complementary.}
As we analyze in Appendix~\ref{app:adaln_intuition}, AdaLN and looping affect visual computation at different granularities. Intuitively, AdaLN provides question-dependent control within each block and feature channel, modifying both the inputs and residual contributions of visual transformations. Also, AdaLN's scaling of a residual update can locally approximate the first-order effect of repeating the transformation (Appendix~\ref{app:adaln_depth}), suggesting layer-wise control over effective depth. Meanwhile, looping explicitly repeats a span of blocks, recomputing their updates as the visual state evolves. Together, they combine fine-grained adaptation with successive applications of the shared computation, and our experiments show benefits from both.

\subsection{Connecting Visual Iteration to Language}

\label{sec:interfaces}
\begin{figure}[t]
  \centering
  \includegraphics[width=0.9\linewidth]{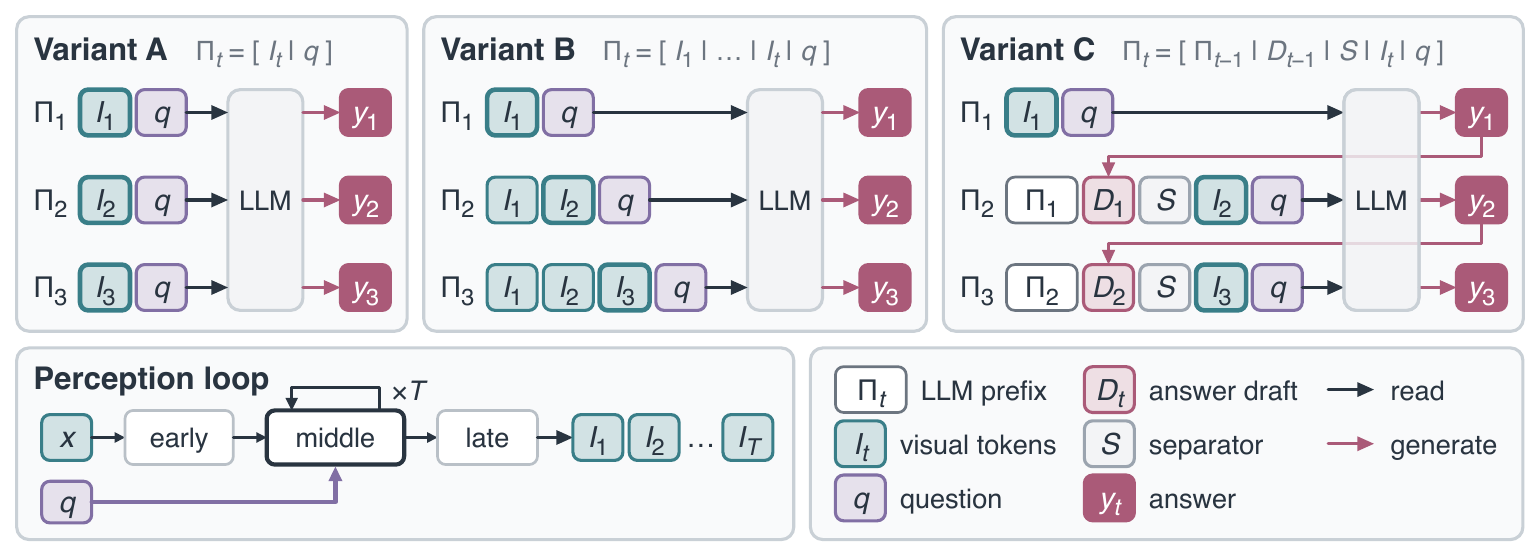}
  \caption{\textbf{Connecting visual iteration to language.}
  The perception loop produces visual tokens $I_t$ across iterations, which are passed to the LLM by concatenation $[\ \cdot\ \vert\ \cdot\ ]$ with question $q$. \textbf{A} generates intermediate and final answers $y_t$ using each $I_t$ in isolation, and \textbf{B} does so using the entire visual history $I_1 \ldots I_t$. \textbf{C} interleaves each $I_t$ with intermediate answers using separator $S$ to distinguish rounds. In all cases, the stability of intermediate answers $y_t$ can be used for adaptive halting.}
  \label{fig:variants}
\end{figure}

We consider three ways for the language model to use the visual tokens produced across iterations (Figure~\ref{fig:variants}). These interfaces differ in whether earlier visual states and language predictions are retained in the context. For variant $v\in\{A,B,C\}$, let $\Pi_t^{(v)}$ denote the language-model prefix at iteration $t$. Here, $q$ denotes the question embeddings, and $[\,\cdot\mid\cdot\,]$ denotes sequence concatenation.

\paragraph{Variant A: current visual state.}
A straightforward strategy is to provide only the visual tokens from the current iteration:
\begin{equation*}
  \Pi_t^{(A)} = [I_t\mid q].
\end{equation*}
This keeps the visual context fixed in length as loop depth increases. Information from earlier visual iterations reaches the language model through the recurrent state $z_t$.

\paragraph{Variant B: accumulated visual states.}
We can give the language model direct access to earlier visual states by concatenating their visual tokens:
\begin{equation*}
  \Pi_t^{(B)} = [I_1\mid\cdots\mid I_t\mid q].
\end{equation*}
The language model can then use the full visual trajectory to predict the answer, therefore reducing the reliance on the current state. This relaxes the need to preserve earlier information in the recurrent state, at the cost of a visual context that grows linearly with loop depth.

\paragraph{Variant C: interleaved generation.}
Beyond retaining visual states, we can also retain the language model's intermediate predictions. Starting from $\Pi_1^{(C)}=[I_1\mid q]$, the language model greedily generates a length-bounded answer draft $D_{t-1}$ at each $t<T$. We append this draft and the next visual input to the prefix:
\begin{equation*}
  \Pi_{t}^{(C)}
    = [\Pi_{t-1}^{(C)}\mid D_{t-1}\mid S\mid I_{t}\mid q],
\end{equation*}
where $S$ is a fixed separator. Subsequent predictions can thus use both the visual history and earlier answers, with additional context and generation cost. The drafts enter only the language context; visual updates continue to use the original question condition.

\subsection{Deep Supervision for Adaptive Inference}
\label{sec:loss}

\paragraph{Training.}
To enable adaptive inference, we supervise predictions throughout every iteration of the visual loop using the same final-answer target. Given an image--question--answer triple $(x,q,y)$ and a maximum training depth $T_{\mathrm{train}}$, the objective is
\begin{equation}
  \mathcal{L}^{(v)}
    = -\sum_{t=1}^{T_{\mathrm{train}}}
        \frac{w_t}{n_y}
        \log p_G\big(y\mid\Pi_t^{(v)}\big),
  \label{eq:deep-supervision}
\end{equation}
where $n_y=|y|$, $w_t>0$ and $\sum_t w_t=1$. We assign larger weights to later depths and compute the loss only on target-answer tokens using teacher forcing. For Variant C, the drafts in the prefix are generated by the model itself. All variants therefore require only final-answer annotations, without intermediate visual or reasoning targets.

By default, we use deep supervision~\citep{lee2015deeply}, evaluating all depth losses for each example. Alternatively, depth-sampled supervision selects a depth with probability $w_t$ and computes its answer-token loss. This gives an unbiased estimate of the objective at lower cost per update.

\paragraph{Inference.}
We use a simple answer-stability rule for adaptive inference. After each visual iteration, the model greedily generates an answer. We stop when two consecutive answers agree or the budget $T_{\mathrm{test}}$ is exhausted and return the current answer in either case. This rule requires intermediate answer generation for all three variants. AdaLoop also supports fixed-depth inference that runs for $T_{\mathrm{test}}$ iterations. In this setting, Variant A and B generate an answer only at the final depth, while Variant C also generates intermediate drafts. In both settings, $T_{\mathrm{test}}$ can differ from $T_{\mathrm{train}}$. Appendix~\ref{app:inference-settings} provides the details of our inference procedure.

\section{VisRO: A Dataset for Vision-Centric Reasoning}
\label{sec:dataset}

\begin{figure}[t]
  \centering
  \captionsetup{skip=8pt}
  \includegraphics[width=0.9\linewidth]{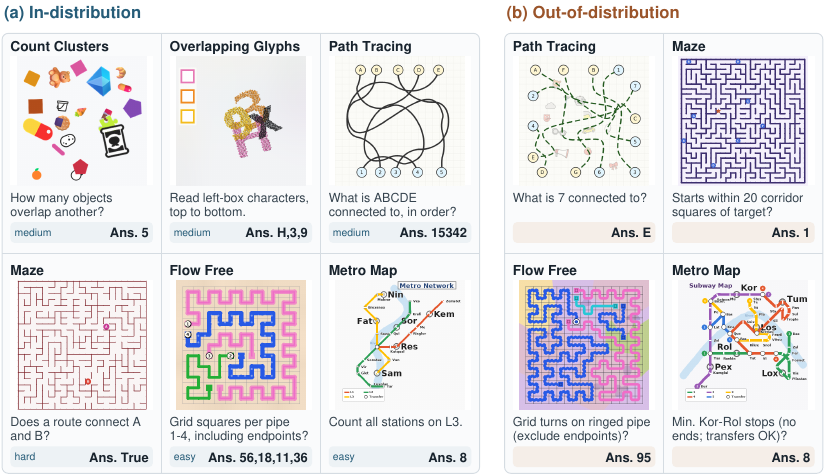}
  \caption{\textbf{VisRO examples.}
  (a) Examples from six core tasks, with level of difficulty shown in blue. (b) Examples from the held-out out-of-distribution (OOD) split that exceed training difficulty (e.g., with larger
  grids) or use held-out generation rules or appearances.
  Each panel shows a $448\times448$ image, abbreviated
  question, and programmatically computed answer.
  Appendix~\ref{app:dataset} has more details.}
  \label{fig:benchmark}
\end{figure}

To study when iterative visual computation helps, we introduce \textbf{VisRO} (Visual Routines), a procedural dataset and generation pipeline for training and evaluating vision-centric reasoning. Inspired by children's visual puzzles and cognitive tests, the collection covers grouping, counting, connectivity, tracing, and recognition under occlusion, with minimal reliance on language priors or external knowledge. We use six tasks from this collection, denoted \textbf{VisRO-Core6}, for our controlled experiments. Figure~\ref{fig:benchmark} shows these tasks and representative out-of-distribution (OOD) examples.

\paragraph{Controlled difficulty and distribution shifts.}

Each core task provides controls over scene complexity, such as object counts, path lengths, overlap, and crossings. We use four difficulty levels for training and in-distribution evaluation. OOD tests include harder instances and shifts in scene generation rules or visual appearance. All VisRO-Core6 images are rendered at $448\times448$. With fixed preprocessing, this lets us vary scene complexity while keeping the per-image visual token budget constant within each backbone.

\paragraph{Generation and evaluation.}
Each generator renders a structured scene and computes ground-truth answers from its underlying representation. Answers are evaluated using task-specific, programmatic scoring rules, without an LLM judge. The underlying scenes can also support spatial supervision, such as object locations and paths through the scene. Our AdaLoop experiments use only image--question--answer triples. Appendix~\ref{app:dataset} describes the broader task collection and task-specific generation procedures.

\section{Experiments}
\label{sec:experiments}

We implement AdaLoop on three pretrained VLMs that differ in both vision
encoder and language backbone: Qwen2.5-VL~\citep{bai2025qwen25vl},
PaliGemma~2~\citep{steiner2024paligemma2}, and LLaVA-1.5~\citep{liu2024improved}.
Our experiments are designed to answer four questions:
(Sec.~\ref{sec:main-results}) Does AdaLoop consistently improve visual reasoning compared to the
non-recurrent base model?
(Sec.~\ref{sec:recurrence-ablation}) Which design choices are critical for effective visual
recurrence?
(Sec.~\ref{sec:adaptive-compute}) When do additional recurrent iterations help at inference time?
(Sec.~\ref{sec:compute-limitations}) Under approximately matched training compute, how does recurrence compare with
non-recurrent alternatives?

\begin{table}[t]
\centering
\caption{\textbf{Results on VisRO-Core6.}
Single-pass question-conditioned AdaLN improves over supervised fine-tuning (SFT), and recurrence (AdaLoop) further improves both in-distribution (ID) and out-of-distribution (OOD) accuracy across all backbones. AdaLoop uses Variant~A with adaptive answer-stability stopping ($p=1$, $T_{\mathrm{test}}=3$). Scores are exact-match accuracy (\%); \emph{Overall} is the macro average over six tasks. \emph{Params}: trainable parameters (M); $\bar{T}$: mean selected stopping depth on the ID set. Gray rows use the same stratified 10\% test subset; API models use their highest available reasoning setting, without tools or a time limit; \emph{tok.}: average generated tokens per completed response, including reasoning tokens. Best overall scores for each backbone are in bold.
}
\label{tab:main-results}
\begingroup
\small
\setlength{\tabcolsep}{2pt}
\renewcommand{\arraystretch}{1.1}
\newlength{\mainresultsoverallwidth}
\settowidth{\mainresultsoverallwidth}{\textbf{Overall}}
\resizebox{0.97\textwidth}{!}{%
\begin{tabular}{lcc*{6}{>{\centering\arraybackslash}m{\mainresultsoverallwidth}}*{2}{>{\columncolor{black!5}\centering\arraybackslash}m{\mainresultsoverallwidth}}}
\toprule
\multirow{2}{*}{Method} & \multirow{2}{*}{Params}
  & \multirow{2}{*}{$\bar{T}$\,/\,tok.}
  & \multicolumn{7}{c}{ID}
  & \multicolumn{1}{c}{OOD} \\
\cmidrule(l{2pt}r{2pt}){4-10}\cmidrule(l{2pt}r{2pt}){11-11}
  & & & Count & Flow & Glyph & Metro & Maze & Path & \textbf{Overall} & \textbf{Overall} \\
\midrule
\multicolumn{11}{@{}l}{\textcolor{gray}{\textit{Closed-source models (zero-shot)}}} \\
\color{gray} Gemini 4 Argon (high) & \color{gray} -- & \color{gray} 6{,}123 & \color{gray} 98.3 & \color{gray} 95.0 & \color{gray} 81.7 & \color{gray} 95.0 & \color{gray} 63.3 & \color{gray} 83.3 & \color{gray} 86.1 & \color{gray} 74.7 \\
\color{gray} GPT-6 Astra (max) & \color{gray} -- & \color{gray} 2{,}892 & \color{gray} 88.3 & \color{gray} 95.0 & \color{gray} 55.0 & \color{gray} 91.7 & \color{gray} 78.3 & \color{gray} 85.0 & \color{gray} 82.2 & \color{gray} 71.4 \\
\color{gray} Claude Opus 5.5 (max) & \color{gray} -- & \color{gray} 31{,}146 & \color{gray} 86.7 & \color{gray} 55.0 & \color{gray} 36.7 & \color{gray} 83.3 & \color{gray} 50.0 & \color{gray} 65.0 & \color{gray} 62.8 & \color{gray} 40.0 \\
\color{gray} Grok 4.6 (high) & \color{gray} -- & \color{gray} 20{,}441 & \color{gray} 36.7 & \color{gray} 0.0 & \color{gray} 36.7 & \color{gray} 33.3 & \color{gray} 31.7 & \color{gray} 36.7 & \color{gray} 29.2 & \color{gray} 15.3 \\
\midrule
\multicolumn{11}{@{}l}{\textcolor{gray}{\textit{Open-source models (zero-shot)}}} \\
\color{gray} GLM-4.6V & \color{gray} -- & \color{gray} 1{,}729 & \color{gray} 33.3 & \color{gray} 1.7 & \color{gray} 23.3 & \color{gray} 18.3 & \color{gray} 13.3 & \color{gray} 10.0 & \color{gray} 16.7 & \color{gray} 5.6 \\
\color{gray} InternVL3.5-241B & \color{gray} -- & \color{gray} 3{,}324 & \color{gray} 21.7 & \color{gray} 1.7 & \color{gray} 28.3 & \color{gray} 28.3 & \color{gray} 26.7 & \color{gray} 13.3 & \color{gray} 20.0 & \color{gray} 11.7 \\
\color{gray} DeepSeek-V4.1-Flash (max) & \color{gray} -- & \color{gray} 20{,}248 & \color{gray} 56.7 & \color{gray} 3.3 & \color{gray} 23.3 & \color{gray} 20.0 & \color{gray} 21.7 & \color{gray} 30.0 & \color{gray} 25.8 & \color{gray} 13.3 \\
\color{gray} Qwen3.5-397B-FP8 & \color{gray} -- & \color{gray} 14{,}476 & \color{gray} 55.0 & \color{gray} 1.7 & \color{gray} 28.3 & \color{gray} 35.0 & \color{gray} 21.7 & \color{gray} 48.3 & \color{gray} 31.7 & \color{gray} 12.2 \\
\midrule
\multicolumn{11}{@{}l}{\textit{Qwen2.5-VL-3B}} \\
\addlinespace[2pt]
Zero-shot & 0 & 1.0 & 10.2 & 0.2 & 1.2 & 12.5 & 31.7 & 16.2 & 12.0 & 9.1 \\
SFT & 66 & 1.0 & 56.8 & 12.0 & 56.0 & 59.3 & 56.2 & 59.3 & 49.9 & 23.9 \\
Single-pass AdaLN & 82 & 1.0 & 61.5 & 23.8 & 84.8 & 60.7 & 56.7 & 72.3 & 60.0 & 30.2 \\
\textbf{AdaLoop} & 82 & 2.2 & 68.5 & 28.0 & 89.2 & 64.0 & 61.8 & 75.8 & \textbf{64.6} & \textbf{32.4} \\
\midrule
\multicolumn{11}{@{}l}{\textit{PaliGemma2-3B}} \\
\addlinespace[2pt]
Zero-shot & 0 & 1.0 & 16.2 & 0.2 & 1.0 & 13.0 & 32.7 & 14.7 & 12.9 & 8.8 \\
SFT & 24 & 1.0 & 74.3 & 26.7 & 80.8 & 67.8 & 70.8 & 55.0 & 62.6 & 31.5 \\
Single-pass AdaLN & 39 & 1.0 & 83.2 & 36.8 & 92.7 & 72.7 & 87.8 & 61.0 & 72.4 & 40.0 \\
\textbf{AdaLoop} & 39 & 2.1 & 84.7 & 40.3 & 93.7 & 75.8 & 93.2 & 63.3 & \textbf{75.2} & \textbf{43.9} \\
\midrule
\multicolumn{11}{@{}l}{\textit{LLaVA-1.5-7B}} \\
\addlinespace[2pt]
Zero-shot & 0 & 1.0 & 6.0 & 0.2 & 0.5 & 13.7 & 29.5 & 12.7 & 10.4 & 8.6 \\
SFT & 341 & 1.0 & 48.3 & 18.0 & 62.8 & 51.7 & 59.5 & 62.3 & 50.4 & 24.8 \\
Single-pass AdaLN & 350 & 1.0 & 52.5 & 29.5 & 82.0 & 53.0 & 63.5 & 65.7 & 57.7 & 29.6 \\
\textbf{AdaLoop} & 350 & 2.2 & 56.8 & 38.7 & 84.7 & 59.0 & 64.2 & 67.3 & \textbf{61.8} & \textbf{31.1} \\
\bottomrule
\end{tabular}
}
\endgroup
\end{table}

\paragraph{Data and evaluation.}
We use the six tasks in VisRO-Core6 (Figure~\ref{fig:benchmark}), with 600k training examples and 3,600 test examples each for in-distribution and OOD evaluation. Each split contains equal numbers of examples from all six tasks. To help retain general visual instruction-following capabilities, we mix the training data with a fixed 100k-example subset of LLaVA-NeXT~\citep{liu2024llavanext}, using its original answer annotations. We report per-task exact-match accuracy and the macro average across tasks, separately for in-distribution and OOD evaluation.

\paragraph{Training and inference.}

Within each backbone, we compare three models, each adding one component to
the previous. The SFT baseline trains the vision--language connector and language-model LoRA adapters~\citep{hu2022lora}, leaving the vision encoder unchanged. Single-pass AdaLN additionally trains the question pooler and modulation generators, and applies the modulated visual blocks only once. AdaLoop trains the same components but applies the loop body blocks repeatedly. All three models use the same training mixture and optimizer-update budget, with corresponding initializations and example order paired by seed. By default, we train AdaLoop through $T_{\mathrm{train}}=3$ and use answer-stability stopping with patience $p=1$ at inference. Details are in Appendix~\ref{app:implementation}.

\Needspace{7\baselineskip}
\subsection{Main Results}

\label{sec:main-results}

Table~\ref{tab:main-results} compares zero-shot inference, supervised fine-tuning (SFT), single-pass AdaLN, and AdaLoop across three pretrained VLMs. AdaLoop uses Variant~A by default. Within each backbone, all instances use the same training data and number of optimizer updates.

\paragraph{Zero-shot evaluation.}
Zero-shot performance varies substantially across models. Qwen2.5-VL, PaliGemma~2, and LLaVA-1.5 achieve $12.0\%$, $12.9\%$, and $10.4\%$ VisRO-ID accuracy, respectively. On a difficulty-stratified 10\% test subset, at their highest available reasoning settings, Gemini~4~Argon~\citep{google2026gemini4argon}, GPT-6 Astra~\citep{openai2026gpt6astra}, Claude Opus~5.5~\citep{anthropic2026opus55}, and Grok~4.6~\citep{xai2026grok46} achieve $86.1\%$, $82.2\%$, $62.8\%$, and $29.2\%$ ID accuracy, respectively. Their average response length, including reasoning tokens, ranges from about 3k tokens for GPT-6 Astra to 31k for Claude Opus~5.5. On the same subset, GLM-4.6V~\citep{glm2025glmv}, InternVL3.5-241B~\citep{wang2025internvl35}, DeepSeek-V4.1-Flash~\citep{deepseekai2026v41flash}, and Qwen3.5-397B-FP8~\citep{qwen2026qwen35} achieve $16.7\%$, $20.0\%$, $25.8\%$, and $31.7\%$ ID accuracy, respectively.

\paragraph{Adapting and reusing visual computation.}
Adding question-conditioned AdaLN improves both ID and OOD accuracy across all three backbones, establishing a strong single-pass baseline. AdaLoop further improves both scores by repeatedly applying the conditioned visual blocks. Relative to single-pass AdaLN, ID accuracy increases by 4.6 percentage points on Qwen2.5-VL, 2.8 on PaliGemma~2, and 4.1 on LLaVA-1.5. These gains are obtained with frozen vision weights and Variant~A, which exposes only the current visual state to the language model. At fixed $T_{\mathrm{test}}$, AdaLoop-A also improves over single-pass AdaLN across all three backbones (Appendix~\ref{app:fixed-depth-results}), without requiring intermediate answer generation for adaptive stopping.

\begin{wraptable}{R}{0.36\textwidth}
\vspace{-1.5em}
\centering
\captionsetup{font=footnotesize,position=top,skip=5pt}
\caption{\textbf{Comparison with existing methods on Qwen2.5-VL.} ID/OOD report accuracy (\%) over VisRO-Core6.}
\label{tab:component-baselines}
\begingroup
\fontsize{8}{9.5}\selectfont
\setlength{\tabcolsep}{7pt}
\renewcommand{\arraystretch}{1.05}
\begin{tabular}{@{}lccc@{}}
\toprule
Method & Params & ID & OOD \\
\midrule
SFT & 66 & 49.9 & 23.9 \\
\quad + GRPO & 66 & 49.5 & 24.2 \\
LIVR & 66 & 47.8 & 22.0 \\
\midrule
AdaLoop & 82 & \textbf{64.6} & \textbf{32.4} \\
\bottomrule
\end{tabular}
\endgroup
\vspace{-0.5em}
\end{wraptable}

\paragraph{\mbox{Comparison with existing methods.}}
Table~\ref{tab:component-baselines} compares AdaLoop with Qwen implementations of SFT+GRPO~\citep{shao2024deepseekmath} and LIVR~\citep{li2026latent}, both using final-answer supervision and matched or more training iterations.
Relative to SFT, GRPO improves OOD accuracy by 0.3 points but reduces ID accuracy, while LIVR falls below SFT on both splits. AdaLoop improves both scores and outperforms these baselines under the evaluated configurations.
Appendix~\ref{app:baseline-implementations} details the baseline implementations.

\Needspace{6\baselineskip}
\subsection{Designing useful visual recurrence}
\label{sec:recurrence-ablation}

We study the language interface, visual modulation, and loop placement in AdaLoop. The ablations use Qwen2.5-VL with frozen vision weights, with training data and optimizer updates matched within each comparison. We use Variant~A except when comparing language interfaces.

\begin{table}[!t]
  \centering
  \captionsetup{position=top,skip=6pt}
  \captionsetup[subtable]{position=bottom,font=footnotesize,labelfont=normalfont,skip=5pt}
  \caption{\textbf{Qwen2.5-VL ablations.} Notation as in Table~\ref{tab:main-results}; best results in bold. (a) All interfaces improve
over single-pass AdaLN, and Variant A performs best. (b) With no modulation, looping gains little over a single pass;  modulation with question conditioning performs better at the same parameter count.}
  \label{tab:design-ablations}
  \newsavebox{\designinterfacesbox}
  \newsavebox{\designcomponentsbox}
  \newlength{\designablationheight}
  \sbox{\designinterfacesbox}{\begingroup%
\fontsize{8}{9.5}\selectfont%
\setlength{\tabcolsep}{4pt}%
\renewcommand{\arraystretch}{1.1}%
\begin{tabular}{lccc}
\toprule
Method & Params & ID & OOD \\
\midrule
Single-pass AdaLN & 82 & 60.0 & 30.2 \\
\midrule
AdaLoop-A & 82 & \textbf{64.6} & \textbf{32.4} \\
AdaLoop-B & 82 & 62.1 & 31.1 \\
AdaLoop-C & 82 & 63.0 & 32.0 \\
\bottomrule
\end{tabular}%
\endgroup%
}
  \sbox{\designcomponentsbox}{\begingroup%
\fontsize{8}{9.5}\selectfont%
\setlength{\tabcolsep}{3.5pt}%
\renewcommand{\arraystretch}{1.1}%
\begin{tabular}{lccccc}
\toprule
\multirow{2}{*}{Modulation} & \multirow{2}{*}{Params} & \multicolumn{2}{c}{Single pass} & \multicolumn{2}{c}{Loop} \\
\cmidrule(lr){3-4}\cmidrule(lr){5-6}
& & ID & OOD & ID & OOD \\
\midrule
None & 66 & 49.9 & 23.9 & 50.9 & 23.8 \\
AdaLN (constant) & 82 & 57.8 & 29.1 & 60.9 & 30.0 \\
AdaLN (question) & 82 & 60.0 & 30.2 & \textbf{64.6} & \textbf{32.4} \\
\bottomrule
\end{tabular}%
\endgroup%
}
  \setlength{\designablationheight}{%
    \dimexpr\ht\designcomponentsbox+\dp\designcomponentsbox\relax}
  \ifdim\dimexpr\ht\designinterfacesbox+\dp\designinterfacesbox\relax>\designablationheight
    \setlength{\designablationheight}{%
      \dimexpr\ht\designinterfacesbox+\dp\designinterfacesbox\relax}
  \fi
  \begin{subtable}[t]{0.48\linewidth}
    \centering
    \begin{minipage}[t][\designablationheight][t]{\linewidth}
      \centering\vspace{0pt}\usebox{\designinterfacesbox}
    \end{minipage}
    \caption{\textbf{Language interfaces.}
    A: current state; B: visual history; C: visual history and
    answer drafts (Fig.~\ref{fig:variants}).}
    \label{tab:variant-comparison}
  \end{subtable}\hfill%
  \begin{subtable}[t]{0.48\linewidth}
    \centering
    \begin{minipage}[t][\designablationheight][t]{\linewidth}
      \centering\vspace{0pt}\usebox{\designcomponentsbox}
    \end{minipage}
    \caption{\textbf{Modulation and recurrence.}
    \emph{Constant} uses a fixed null question. Looping uses variant A.}
    \label{tab:component-ablations}
  \end{subtable}
\end{table}

\begin{figure}[t]
  \centering
  \captionsetup[subfigure]{font=footnotesize,labelfont=bf,skip=3pt}
  \newsavebox{\adaptiveinferencetablebox}
  \newsavebox{\adaptiveinferenceplotbox}
  \newlength{\adaptiveinferenceheight}
  \sbox{\adaptiveinferencetablebox}{%
    \begin{minipage}{0.52\linewidth}
      \centering
      \begingroup
\fontsize{8}{9.5}\selectfont
\setlength{\tabcolsep}{1pt}
\renewcommand{\arraystretch}{1.08}
\newcommand{\gainheading}{Rel. gain}
\begin{tabular*}{\linewidth}{@{\extracolsep{\fill}}lcccccc@{}}
  \toprule
  \multirow{2}{*}[-0.8ex]{Difficulty} & \multicolumn{2}{c}{Qwen2.5-VL} & \multicolumn{2}{c}{PaliGemma~2} & \multicolumn{2}{c}{LLaVA-1.5} \\
  \cmidrule(l{1pt}r{1pt}){2-3}\cmidrule(l{1pt}r{1pt}){4-5}\cmidrule(l{1pt}r{1pt}){6-7}
  & \gainheading & $\bar{T}$ & \gainheading & $\bar{T}$ & \gainheading & $\bar{T}$ \\
  \midrule
  Easy   & +3.33 & 2.11 & +0.87 & 2.06 & +3.05 & 2.11 \\
  Medium & +5.87 & 2.17 & +0.00 & 2.09 & $-$0.17 & 2.15 \\
  Hard   & +4.09 & 2.21 & +1.93 & 2.14 & +4.87 & 2.21 \\
  XHard  & +4.46 & 2.26 & +5.43 & 2.21 & +7.85 & 2.28 \\
  \midrule
  OOD    & +6.39 & 2.30 & +4.49 & 2.23 & +6.57 & 2.29 \\
  \bottomrule
\end{tabular*}
\endgroup

    \end{minipage}}
  \sbox{\adaptiveinferenceplotbox}{%
    \includegraphics[width=0.42\linewidth]{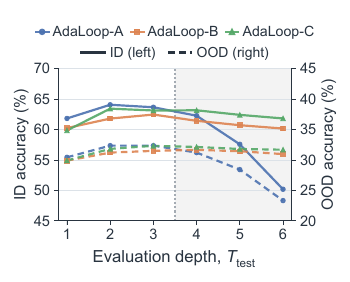}}
  \setlength{\adaptiveinferenceheight}{%
    \dimexpr\ht\adaptiveinferenceplotbox+\dp\adaptiveinferenceplotbox\relax}
  \ifdim\dimexpr\ht\adaptiveinferencetablebox+\dp\adaptiveinferencetablebox\relax>\adaptiveinferenceheight
    \setlength{\adaptiveinferenceheight}{%
      \dimexpr\ht\adaptiveinferencetablebox+\dp\adaptiveinferencetablebox\relax}
  \fi
  \begin{subfigure}[t]{0.52\linewidth}
    \centering
    \begin{minipage}[c][\adaptiveinferenceheight][c]{\linewidth}
      \centering
      \usebox{\adaptiveinferencetablebox}
    \end{minipage}
    \caption{\textbf{Adaptive computation.}
    Relative accuracy gain (\%) of answer-stability stopping over one visual iteration of the same checkpoint, and mean stopping depth $\bar{T}$.}
    \label{fig:adaptive-difficulty-table}
  \end{subfigure}\hfill
  \begin{subfigure}[t]{0.46\linewidth}
    \centering
    \begin{minipage}[c][\adaptiveinferenceheight][c]{\linewidth}
      \centering
      \usebox{\adaptiveinferenceplotbox}
    \end{minipage}
    \caption{\textbf{Depth extrapolation.}
    ID accuracy (solid, left axis) and OOD accuracy (dashed, right axis) on Qwen2.5-VL. Shading marks $T_{\text{test}} > T_{\text{train}}$.}
    \label{fig:abc-loop-depth}
  \end{subfigure}
  \caption{\textbf{Inference-time visual computation.} All models are trained  with $T_{\mathrm{train}}=3$. (a) Mean stopping depth increases with difficulty
on every backbone. (b) Beyond the training depth, Variants B and C remain largely stable, whereas A degrades.}
  \label{fig:adaptive-inference}
\end{figure}

\paragraph{Language interfaces.}
All three interfaces improve both ID and OOD accuracy over single-pass AdaLN, and reading only the current state (A) performs best (Table~\ref{tab:variant-comparison}), so explicitly retaining earlier visual states or answer drafts in the language context is not required to benefit from visual recurrence. Retaining them does help beyond the training depth, however: B and C degrade less than A when run for more iterations (Figure~\ref{fig:abc-loop-depth}). One possible explanation is that the retained history lets the language model draw on earlier visual states, which are unaffected when later iterations drift beyond the trained regime.

\paragraph{Modulation and recurrence.}
Table~\ref{tab:component-ablations} separates the contributions of modulation, question conditioning, and recurrence. Without modulation, looping raises ID accuracy only from 49.9\% to 50.9\%.
Adding question-independent AdaLN improves the loop to 60.9\%, partly through its 16M additional trainable parameters. Conditioning the same modules on the question further raises accuracy to 64.6\%, without adding parameters. With question conditioning held fixed, looping improves over single-pass AdaLN by 4.6 ID and 2.2 OOD percentage points. Thus, question information and recurrent computation each contribute beyond the addition of trainable modulation parameters.

\begin{figure}[t]
  \centering
  \captionsetup[subfigure]{font=footnotesize,labelfont=bf,skip=4pt,justification=raggedright,singlelinecheck=false}
  \newsavebox{\spanselectionckabox}
  \sbox{\spanselectionckabox}{%
    \includegraphics[width=0.3\linewidth]{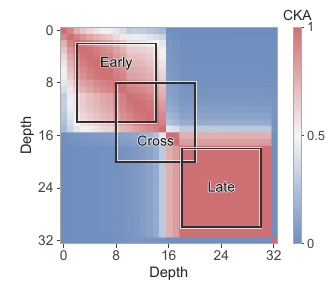}}
  \begin{subfigure}[t]{0.33\linewidth}
    \vspace{0pt}
    \centering
    \usebox{\spanselectionckabox}
    \caption{\textbf{Encoder representation similarity.} Linear CKA across depths $k$, and three candidate spans.}
    \label{fig:cka-heatmaps}
  \end{subfigure}\hfill
  \begin{minipage}[t]{0.65\linewidth}
    \vspace{0pt}
  \begin{subfigure}[t]{\linewidth}
    \vspace{0pt}
    \centering
    \begingroup
\fontsize{8}{9.5}\selectfont
\setlength{\tabcolsep}{1.4pt}
\renewcommand{\arraystretch}{1.08}
\begin{tabular*}{\linewidth}{@{\extracolsep{\fill}}lccccccc@{}}
\toprule
Loop span & Count & Flow & Glyph & Metro & Maze & Path & Overall \\
\midrule
Early $[2,14)$ & \textbf{11.38} & \textbf{17.48} & \textbf{5.11} & \textbf{5.49} & 9.12 & 4.84 & \textbf{7.64} \\
Cross $[8,20)$ & 4.52 & 8.23 & 3.87 & 1.63 & \textbf{22.70} & $-$0.47 & 6.05 \\
Late $[18,30)$ & 6.70 & 2.63 & 2.28 & 2.58 & $-$0.56 & \textbf{5.57} & 3.33 \\
\bottomrule
\end{tabular*}
\endgroup

    \caption{\textbf{Gains from recurrence.} For each loop span [$k_\text{start},k_\text{end}$), relative VisRO-ID accuracy gains (\%) over a single-pass AdaLN, under answer-stability halting.}
    \label{fig:qwen-span-gains}
  \end{subfigure}

\caption{\textbf{Qwen2.5-VL loop placement.} (a) Three candidate loop spans that are contained within, or that cross, encoder spans having high CKA similarity. (b) Recurrence improves overall accuracy for all spans. Gains vary by task for each loop span.}
  \label{fig:span-selection}
  \end{minipage}

\end{figure}

\paragraph{Loop placement.}
Figure~\ref{fig:cka-heatmaps} shows pretrained visual representation similarity measured by centered kernel alignment (CKA)~\citep{kornblith2019similarity}. On Qwen2.5-VL, we compare three equal-length spans: early $[2,14)$, cross-band $[8,20)$, and late $[18,30)$ (Figure~\ref{fig:qwen-span-gains}). All three improve overall ID accuracy over their corresponding single-pass AdaLN controls.
The early span gives the largest overall relative gain and leads on four tasks, while the cross-band and late spans give the largest relative gains on Maze and Path, respectively. The benefits of recurrence therefore depend on both placement and task; useful recurrence also occurs in a span crossing the transition between high-CKA bands. CKA measurement details are provided in Appendix~\ref{sec:cka}.

\Needspace{18\baselineskip}
\subsection{Inference-time Visual Computation}

\label{sec:adaptive-compute}
\begin{wrapfigure}[24]{R}{0.45\linewidth}
  \vspace{-3.0em}
  \centering
  \captionsetup{font=footnotesize,skip=4pt}
  \includegraphics[width=\linewidth]{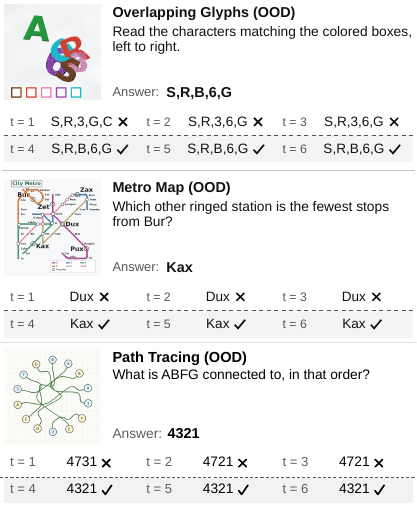}
  \caption{\textbf{Selected answer trajectories beyond the training depth} (Qwen2.5-VL, Variant C). At each depth $t$, the model answers from the visual history and its earlier drafts. In these selected examples, the shaded iterations ($t > T_{\mathrm{train}}$) correct the answer and retain it.}
  \label{fig:self-correction-trajectories}
\end{wrapfigure}

Recurrence and deep supervision enable AdaLoop to use adaptive computation at inference time. We examine how this benefit varies with problem difficulty and whether useful updates continue beyond the depths seen during training.

\paragraph{Adaptive computation within trained depths.}
Using the same checkpoint ($T_{\mathrm{train}}=3$), we compare adaptive stopping ($p=1$, $T_{\mathrm{test}}=3$) with stopping after the first iteration. As shown in Figure~\ref{fig:adaptive-difficulty-table}, adaptive stopping improves accuracy in most difficulty groups across all three backbones. Its mean stopping depth also increases with difficulty on every backbone, although neither the model nor the stopping rule sees difficulty labels. This shows that adaptive stopping using answer stability allocates more iterations to harder inputs.

\paragraph{Behavior beyond trained depths.}
Figure~\ref{fig:abc-loop-depth} shows that Variants B and C largely maintain accuracy beyond $T_{\mathrm{train}}=3$, whereas A degrades more as depth increases. Although average
accuracy stays roughly flat, this stability leaves room for later iterations
to correct individual examples: Figure~\ref{fig:self-correction-trajectories} (Variant C) illustrates such cases, where predictions on Glyphs, Metro Map, and Path Tracing become correct at depth four and remain correct through depth six.

\Needspace{4\baselineskip}
\subsection{Computational Trade-offs}

\label{sec:compute-limitations}
AdaLoop achieves better performance at matched optimizer updates (Sec.~\ref{sec:main-results}), but looping incurs higher training and inference costs than a parameter-matched single-pass model. We quantify these costs on Qwen2.5-VL and compare models under approximately matched training compute.

\paragraph{Inference cost.}
On VisRO-ID and with fixed-depth inference at $T_{\mathrm{test}}=3$, AdaLoop generates an answer only after the final visual iteration, using an estimated $1.3\times$ the inference FLOPs of single-pass AdaLN.
Answer-stability stopping with the same depth limit has an estimated cost of $2.2\times$ on average because it also generates intermediate answers. Appendix~\ref{app:abc-cost} describes the accounting.

\begin{figure}[t]
  \centering
  \captionsetup[subfigure]{font=footnotesize,labelfont=bf,skip=3pt}
  \newsavebox{\trainingradarbox}
  \newlength{\trainingcomparisonheight}
  \sbox{\trainingradarbox}{%
    \includegraphics[width=0.33\linewidth]{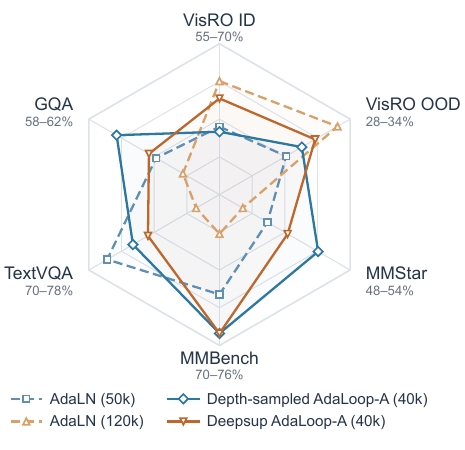}}
  \setlength{\trainingcomparisonheight}{%
    \dimexpr\ht\trainingradarbox+\dp\trainingradarbox\relax}
  \begin{subfigure}[t]{0.34\linewidth}
    \vspace{0pt}
    \centering
    \usebox{\trainingradarbox}
    \caption{\textbf{Performance profiles.}
    Colors pair approximately matched training budgets; axes ranges differ as labeled.}
    \label{fig:training-tradeoffs-radar}
  \end{subfigure}\hfill
  \begin{subfigure}[t]{0.65\linewidth}
    \vspace{0pt}
    \centering
    \begin{minipage}[c][\trainingcomparisonheight][c]{\linewidth}
      \centering
      \begingroup
\fontsize{8}{9}\selectfont
\setlength{\tabcolsep}{1.4pt}
\renewcommand{\arraystretch}{1.15}
\resizebox{0.96\textwidth}{!}{%
\begin{tabular}{@{}lcccccc@{}}
\toprule
\multirow{2}{*}{Method} & Train & VisRO & \multirow{2}{*}{MMStar} & \multirow{2}{*}{MMBench} & \multirow{2}{*}{GQA} & \multirow{2}{*}{TextVQA} \\
 & FLOPs & (ID+OOD) & & & & \\
\midrule
AdaLN (40k) & $1.00\times$ & 45.10 & 51.07 & 75.43 & 60.28 & 76.28 \\
\midrule
AdaLN (50k) & $1.25\times$ & 46.40 & 50.20 & 73.97 & 59.94 & 76.88 \\
AdaLoop (sampled, 40k) & $1.32\times$ & 46.51 & 52.53 & 75.52 & 61.15 & 75.30 \\
\midrule
AdaLN (120k) & $3.00\times$ & 49.85 & 49.07 & 71.56 & 59.12 & 71.43 \\
AdaLoop (deep sup., 40k) & $2.86\times$ & 48.47 & 51.13 & 75.52 & 60.16 & 74.38 \\
\bottomrule
\end{tabular}%
}\par
\endgroup

    \end{minipage}\par
    \captionsetup{margin={0.02\linewidth,0.02\linewidth}}
    \caption{\textbf{Training cost and accuracy.}
    \emph{Sampled} supervises one sampled depth per example; \emph{deep sup.} supervises all depths.
    Training FLOPs are relative to single-pass AdaLN at 40k updates (Appendix~\ref{app:abc-cost}). VisRO averages ID and OOD accuracy (\%).}
    \label{fig:training-budget}
  \end{subfigure}
  \caption{\textbf{Qwen2.5-VL training compute allocation.}
  (a) Radar visualization of numbers in Table (b). At approximately matched training FLOPs, depth-sampled AdaLoop matches the 50k-update single-pass control on VisRO and improves three of four VQA benchmarks. Deeply supervised AdaLoop is slightly below the 120k-update control on VisRO but outperforms it on all four general VQA benchmarks, using one third as many optimizer updates.}
  \label{fig:training-tradeoffs}
\end{figure}

\paragraph{Training cost.}
At $T_{\mathrm{train}}=3$, deep supervision at every depth costs approximately $2.9\times$ the training FLOPs per update of single-pass AdaLN.
Sampling one supervised depth according to the normalized loss weights reduces the estimated cost to approximately $1.3\times$, while preserving the weighted objective in expectation.
We therefore compare depth-sampled and deeply supervised AdaLoop, each trained
for 40k updates, with single-pass AdaLN trained for 50k and 120k updates,
respectively; total training FLOPs match to within 6\% (Figure~\ref{fig:training-budget}). AdaLoop thus uses 20\% and 67\% fewer optimizer updates, and correspondingly fewer passes over the training data.

\paragraph{Performance at approximately matched training compute.}
Besides VisRO, we also evaluate on GQA~\citep{hudson2019gqa}, MMBench~\citep{liu2024mmbench}, MMStar~\citep{chen2024we}, and TextVQA~\citep{singh2019towards}. These benchmarks measure retention of general VQA capabilities acquired by the backbones during their original training, and our post-training mixture includes a LLaVA-NeXT subset to limit forgetting (Section~\ref{sec:experiments}).

At the lower budget (Figure~\ref{fig:training-tradeoffs}), depth-sampled AdaLoop matches the 50k-update control on VisRO and is higher on three general VQA benchmarks. At the higher budget, deeply supervised AdaLoop is slightly below the 120k-update control on VisRO but outperforms it on all four VQA benchmarks. Extending single-pass AdaLN training from 40k to 120k updates lowers all general benchmarks (e.g., MMBench from 75.4\% to 71.6\%), consistent with forgetting. These results suggest that recurrence offers a different way to spend training compute: more computation per example rather than more updates over the same data. At approximately matched FLOPs, AdaLoop achieves VisRO accuracy competitive with extended single-pass training while retaining better general capabilities.

\section{Conclusion}

We introduced AdaLoop, a framework for question-conditioned visual recurrence, and VisRO, a procedural dataset and generation pipeline for vision-centric reasoning. Across three VLM families, question conditioning and recurrence provide complementary gains. With training data and optimizer updates matched, AdaLoop improves over both supervised fine-tuning and question-conditioned single-pass controls. These gains are obtained with the pretrained vision weights frozen. Our findings suggest that a pretrained vision encoder need not be a fixed front end: its existing visual transformations can be adapted and reused for iterative reasoning. We hope this work encourages further study of how vision modules can contribute to multimodal reasoning.

\FloatBarrier

\let\normalsize\arxivSavedNormalsize
\parskip=\arxivSavedparskip
\textfloatsep=\arxivSavedtextfloatsep
\floatsep=\arxivSavedfloatsep
\abovedisplayskip=\arxivSavedabovedisplayskip
\belowdisplayskip=\arxivSavedbelowdisplayskip
\abovedisplayshortskip=\arxivSavedabovedisplayshortskip
\belowdisplayshortskip=\arxivSavedbelowdisplayshortskip
\captionsetup[figure]{skip=10pt}

\subsubsection*{Acknowledgments}
The authors thank Kaiming He for insightful discussions on the AdaLoop framework and Jianbo Shi for proofreading the manuscript. We also thank Keya Hu and Qiao Sun for helpful discussions on TPU infrastructure. We thank Google TPU Research Cloud (TRC) for granting us access to TPUs.

{
\bibliographystyle{iclr2027_conference}
\bibliography{main}
}

\appendix
\clearpage

\section{AdaLN as Task-Conditioned Visual Compute Allocation}
\label{app:adaln_intuition}

This section gives an intuitive account of why question-conditioned adaptive
layer normalization (AdaLN) is useful when adapting a pretrained vision
encoder.  The purpose is not to establish an exact equivalence between feature
modulation and recurrent computation.  Rather, a local analysis shows that
AdaLN provides a fine-grained, first-order mechanism for reallocating the
visual transformations already present in the encoder, while recurrence
provides the complementary ability to recompute those transformations at an
updated visual state. Throughout, ``compute allocation'' refers to the
relative influence of transformations already available in the encoder rather
than to arithmetic: modulation changes which pretrained updates dominate
without changing the FLOPs of a single pass.  Recurrence is what increases
executed depth.

\subsection{From AdaLN to a modulated residual update}
\label{app:adaln_residual}

For clarity, consider one pre-normalized residual branch at layer $\ell$ and
loop iteration $t$.  Let $x_{\ell,t}\in\mathbb{R}^{N\times d}$ denote $N$
visual tokens with channel dimension $d$, and let $q$ denote the question
representation.  We write our modulation abstractly as
\begin{align}
    z_{\ell,t} &= \operatorname{LN}(x_{\ell,t}),
    \label{eq:adaln_ln}\\
    \widetilde z_{\ell,t}
        &= \bigl(\mathbf{1}+\gamma_{\ell}(q)\bigr)\odot z_{\ell,t}
           + \beta_{\ell}(q),
    \label{eq:adaln_def}\\
    x^{+}_{\ell,t}
        &= x_{\ell,t}
           +\bigl(\mathbf{1}+\alpha_{\ell}(q)\bigr)\odot
            F_{\ell}(\widetilde z_{\ell,t}),
    \label{eq:adaln_block}
\end{align}
where $F_{\ell}$ is the update produced by the transformer layer, and $\alpha_{\ell},\beta_{\ell},\gamma_{\ell}\in\mathbb{R}^{d}$ are the
residual gate, shift, and scale of Eq.~\eqref{eq:adaln}, broadcast over visual
tokens. We suppress the sublayer superscript $f$, since the same notation
applies separately to the attention and MLP sub-layers of a transformer block.
Conditioning on $q$ allows the same
pretrained block to emphasize different channels for different questions.  Unless
written as a function of $t$, the modulation is taken to be reused across loop iterations, as in our Qwen and PaliGemma models. LLaVA AdaLoop additionally adds a learned iteration embedding to the condition (Appendix~\ref{app:question-conditioning}), so its modulation also depends on $t$.

Let $u^0_{\ell,t}=F_{\ell}(z_{\ell,t})$ be the update produced by the
unmodulated sub-layer and define
\(
\delta z_{\ell,t}=\gamma_{\ell}\odot z_{\ell,t}+\beta_{\ell},
\) so that $\widetilde z_{\ell,t}=z_{\ell,t}+\delta z_{\ell,t}$.  A first-order
expansion around the pretrained computation gives
\begin{equation}
    x^{+}_{\ell,t}
    = x_{\ell,t}
      +(\mathbf{1}+\alpha_{\ell})\odot
       \Bigl(u^0_{\ell,t}
        +J_{F_{\ell}}(z_{\ell,t})
         \bigl[\gamma_{\ell}\odot z_{\ell,t}+\beta_{\ell}\bigr]\Bigr)
      +\mathcal{O}(\|\delta z_{\ell,t}\|^2).
    \label{eq:adaln_local_expansion}
\end{equation}
Equation~\eqref{eq:adaln_local_expansion} makes the computational role of
AdaLN explicit: it does not create a new visual operator from scratch, but
steers the update generated by a pretrained operator using a low-dimensional,
question-dependent intervention.  The scale
$\gamma_{\ell}$ is multiplicative in the current visual activity, the gate
$\alpha_{\ell}$ rescales the resulting update channel by channel, and the
shift $\beta_{\ell}$ injects a channel-wise bias shared across visual tokens.  Thus, the two multiplicative terms preserve the image- and token-dependent
structure of $z_{\ell,t}$ and can selectively amplify or suppress
transformations that are already active; the additive term does not. This qualitative
distinction does not by itself prove that scaling must be more effective.

\subsection{A first-order connection to effective depth}
\label{app:adaln_depth}

Throughout this section we write the visual state of a block as a single vector
$x \in \mathbb{R}^{Nd}$, stacking its $N$ tokens, and $F\colon\mathbb{R}^{Nd} \to
\mathbb{R}^{Nd}$ for the residual branch acting on it. We allow $F$ to mix tokens,
as attention does, so its Jacobian carries both channel and cross-token dependence.
A channel-wise quantity $v \in \mathbb{R}^d$ acts on the state as the matrix
$\Gamma_v = I_N \otimes \operatorname{Diag}(v) \in \mathbb{R}^{Nd \times Nd}$, with
$\operatorname{Diag}$ making the action channel-wise and $I_N\otimes$ applying it
identically to every token. Such matrices compose as
$\Gamma_u \Gamma_v = \Gamma_{u \odot v}$.

To isolate the scale intuition, consider a surrogate model where the question
modulates the block only by rescaling its update:
\begin{equation}
x^+ \;=\; x + \Gamma_{g}\, F(x),
\label{eq:gain_surrogate}
\end{equation}
with question- and depth-dependent channel-wise gain $g(q,t) \in
\mathbb{R}^{d}$. This section and the next analyze which effects of recurrence---repeated application of $F$---a single gain-modulated pass can reproduce.

Equation~\eqref{eq:gain_surrogate} is an identity for AdaLN's gate path because the
residual gate contributes $x^+ = x + \Gamma_{\mathbf{1}+\alpha} F(x)$. Whether the pre-branch scale
$\gamma$ and shift $\beta$ can be absorbed into the same gain is less immediate. The scale
reduces to a gain in a restricted case: if the branch acts on the normalized input $z$
of \eqref{eq:adaln_ln} by a shared channel-wise scaling, $F(z) = \Gamma_\lambda z$ with
$\lambda \in \mathbb{R}^d$, then $\Gamma_\lambda \Gamma_{\mathbf{1}+\gamma} = \Gamma_{\mathbf{1}+\gamma}
\Gamma_\lambda$, recovering \eqref{eq:gain_surrogate} with $\Gamma_g =
\Gamma_{\mathbf{1}+\alpha} \Gamma_{\mathbf{1}+\gamma}$. Channel mixing, token mixing, and nonlinearities, including
the attention softmax, break this commutation, so in general the scale matches a gain
only in a neighborhood of the current state, and it changes as that state evolves. The
shift matches no gain at all, entering by \eqref{eq:adaln_local_expansion} as a bias
injection rather than a rescaling, we omit it to isolate the effect of scaling, so the analysis
below concerns the multiplicative part of AdaLN only.
We use
\eqref{eq:gain_surrogate} as an interpretive model of the combined effect, not as an
identity for the full nonlinear block.

Residual networks admit the familiar interpretation of an Euler discretization
of a continuous transformation~\citep{chen2018neural}.  Introduce a nominal
step size $\epsilon$, equal to $1$ in a pretrained block, and define
\begin{equation}
    R_{\epsilon}(x)=x+\epsilon F(x).
\end{equation}
For a fixed integer $n$, applying this update $n$ times and expanding around
the initial state yields
\begin{equation}
    R_{\epsilon}^{n}(x)
    =x+n\epsilon F(x)
     +\binom{n}{2}\epsilon^2J_F(x)F(x)
     +\mathcal{O}_{n}(\epsilon^3),
    \label{eq:recurrent_expansion}
\end{equation}
where the subscript on $\mathcal{O}_{n}$ indicates a constant that grows with
$n$. (The expansion says which terms recurrence contributes, not how large
they are.)

By contrast, a single scalar-gated update, with $c\in\mathbb{R}$, is
\begin{equation}
    \widehat R_{\epsilon,c}(x)
    =x+\epsilon c F(x).
    \label{eq:gain_one_step}
\end{equation}
Setting $c=n$ reproduces the first-order displacement in
Eq.~\eqref{eq:recurrent_expansion} exactly, so the two leading terms in that expression divide
between the two mechanisms. The gain accounts for the displacement, meaning how far
the representation moves along the currently available update direction, with
$c$ free to be fractional. Recurrence accounts for the second-order term $J_F(x)F(x)$, which exists because $F$ is reevaluated after the state moves. Residual scaling therefore allocates effective depth, and recurrence
supplies what follows from recomputing the direction.

The linear case makes this distinction especially transparent, because there the $n$-step update is exact instead of an expansion. If $F(x)=Wx$ with $W \in \mathbb{R}^{Nd \times Nd}$, then
\begin{equation}
    R_{\epsilon}^{n}(x)=(I+\epsilon W)^n x,
    \qquad
    \widehat R_{\epsilon,\Gamma_g}(x)
        =(I+\epsilon\Gamma_g W)x.
    \label{eq:linear_loop_vs_gain}
\end{equation}
Suppose $W$ shares the structure of $\Gamma_g$, acting identically and
diagonally on every token: $W=\Gamma_\lambda$ with
$\lambda=(\lambda_1,\ldots,\lambda_d)\in\mathbb{R}^d$.
A channel-wise gain can then match the $n$-step linear transformation exactly within this residual-gain surrogate:
\begin{equation}
    g_i^{\star}(n)
    =\frac{(1+\epsilon\lambda_i)^n-1}
           {\epsilon\lambda_i},
    \qquad
    g_i^{\star}(n)=n\ \text{when }\lambda_i=0.
    \label{eq:optimal_channel_gain}
\end{equation}
For small $\epsilon\lambda_i$, this reduces to
\(
g_i^{\star}(n)=n+\binom{n}{2}\epsilon\lambda_i+\cdots,
\)
so the gain that reproduces $n$ iterations is, to leading order, $n$ itself. Moreover $n$ enters Eq.~\eqref{eq:optimal_channel_gain} only as an exponent, so
$g_i^{\star}$ is defined for non-integer $n$, and the gain can interpolate
continuously between integer depths and hold a different effective depth in
each channel.  The correspondence holds only while $W$ shares both of
$\Gamma_g$'s factors, however.  If $W$ varies across tokens or mixes them,
as attention does, exact matching would require a gain that does the same,
which the factor $I_N$ forbids; and if $W$ is dense in channels, the target
$(I+\epsilon W)^n$ contains the mixtures $W^2,W^3,\ldots$ that a diagonal gain
cannot produce.

The two obstructions just described are the linear case of a more general
mismatch, which is stronger still when $F$ is nonlinear, because the Jacobian
changes with the evolving state.
The leading discrepancy after scalar first-order matching ($c=n$) is
already visible in Eq.~\eqref{eq:recurrent_expansion}:
\begin{equation}
    R_{\epsilon}^{n}(x)-\widehat R_{\epsilon,n}(x)
    =\binom{n}{2}\epsilon^2J_F(x)F(x)
     +\mathcal{O}_{n}(\epsilon^3).
    \label{eq:leading_mismatch}
\end{equation}
A richer gain absorbs part of this term but not all of it.  Allowing the gain
its own first-order correction, $\Gamma_{n\mathbf{1}+\epsilon\mu}$ with
$\mu\in\mathbb{R}^d$, adds a second-order term
$\epsilon^2\Gamma_{\mu}F(x)$. Matching
$\binom{n}{2}\epsilon^2J_F(x)F(x)$ then requires
$\mu_i[F(x)]_{j,i}=\binom{n}{2}[J_F(x)F(x)]_{j,i}$ at every token $j$ and
channel $i$, which is $Nd$ equations in $d$ unknowns.  A token-shared gain can
therefore absorb whatever part of $J_F(x)F(x)$ is a token-uniform channel-wise
rescaling of $F(x)$, but not the part that varies across tokens.

The quantity $J_F(x)F(x)$ measures how the appropriate update changes after one
refinement step, which suggests a broader task-level hypothesis.  Recurrence should
be most useful when solving a task that requires successive updates to the internal
relational organization of a scene, such as propagating reachability
through a maze or forming and consolidating visual groups before counting
them.  In such cases, each partial assignment changes where or how the next
update should act, producing computation that cannot generally be reduced to
rescaling a fixed update direction.  By contrast, tasks whose relevant
structure is already captured by a stable feedforward representation may
benefit less from additional visual recurrence.
Whether a task demands iterative state construction depends on how much structure a single pass already resolves, not on the task's label.

\subsection{Why AdaLN and recurrence are complementary}
\label{app:adaln_loop_complementarity}

The analysis above separates two axes that ordinary network depth conflates.
AdaLN provides \emph{allocation}: it selects, per layer and per channel, how
strongly a pretrained transformation contributes for a given question.
Recurrence provides \emph{recomputation}: it reevaluates that transformation at
the state produced by the preceding iteration.  They also differ in
granularity: AdaLN acts per channel and continuously, by
\eqref{eq:optimal_channel_gain}, whereas a loop acts per span and in discrete
steps.

This view also clarifies why a uniform, unconditioned loop can be a poor way to
reuse a pretrained hierarchy.  Different blocks implement transformations at
different levels of abstraction, different questions need not benefit from
repeating all of them equally, and a single loop count imposes the same coarse
allocation on every block in the recurrent span.

AdaLN's conditioning removes that constraint. The frozen blocks and the question-derived modulation are both reused at every iteration, so the loop body is a single shared operator; but because the modulation depends on $q$, that operator becomes a family indexed by the question:
\begin{equation}
    x_{t+1}=\mathcal{L}\bigl(x_t;\{\alpha_\ell(q),\beta_\ell(q),\gamma_\ell(q)\}_{\ell\in\mathcal{B}}\bigr),
    \label{eq:conditioned_loop}
\end{equation}
where $\mathcal{L}$ denotes the shared loop body and $\mathcal{B}$ its set of
blocks.  Training learns the map $q \mapsto \{\alpha_\ell(q),\beta_\ell(q),\gamma_\ell(q)\}$ rather than a single
update rule, so different questions emphasize different feature dimensions of
the same pretrained span.  The updates applied at successive iterations differ because, while $\mathcal{L}$ is fixed across $t$, its input
$x_t$ is not.

\paragraph{Takeaway.}
AdaLN and looping are not competing approximations to the same
operation.  AdaLN selects and reweights the
pretrained visual computation appropriate for a question, which locally
resembles a channel-wise, fractional allocation of effective depth.  Looping
supplies what a single modulated pass cannot: repeated evaluation at the evolving
visual state, including the higher-order term $J_FF$ of \eqref{eq:leading_mismatch} associated with propagation
and feedback.  Their combination realizes task-conditioned iterative
visual computation under final-answer supervision alone.

\section{Implementation Details}
\label{app:implementation}

Table~\ref{tab:backbone-config} lists the model configurations and training hyperparameters. Below, we describe the adapters, training procedure, inference protocol, and baseline implementations. Dataset generation and scoring are described in Appendix~\ref{app:dataset}; FLOP accounting is given in Appendix~\ref{app:abc-cost}. Training and evaluation code for all three backbones is available at \url{https://github.com/xrhan/AdaLoop}, and model checkpoints at \url{https://huggingface.co/Lyy0725/AdaLoop}.

\subsection{Model and Adapter Configurations}
\label{app:question-conditioning}

\begin{table}[!t]
\centering
\captionsetup{position=top}
\caption{\textbf{Configurations and training hyperparameters.} Defaults for the main experiments; exceptions are specified with the corresponding results. Batch sizes are global, and training length and warmup count optimizer updates. $N$ denotes the number of recurrent visual tokens in Qwen. Encoder references: SigLIP So400m/14~\citep{zhai2023siglip,alabdulmohsin2023getting}, CLIP ViT-L/14~\citep{radford2021clip}. $^{\S}$We use our re-trained LLaVA-1.5-7B checkpoint at $448\times448$, with $2\times2$ average pooling of the 1{,}024 CLIP patch tokens, rather than the released 336-px checkpoint.}
\label{tab:backbone-config}
\vspace{5pt}
\begingroup
\small
\setlength{\tabcolsep}{5pt}
\renewcommand{\arraystretch}{1.1}
\begin{tabularx}{\linewidth}{@{}l*{3}{>{\centering\arraybackslash}X}@{}}
\toprule
Setting & Qwen2.5-VL-3B & PaliGemma~2~3B & LLaVA-1.5-7B$^{\S}$ \\
\midrule
Vision encoder & Native ViT & SigLIP So400m/14 & CLIP ViT-L/14 \\
Vision hidden width $d_v$ & 1280 & 1152 & 1024 \\
LM embedding width & 2048 & 2304 & 4096 \\
Image preprocessing & Dynamic & $448\times448$ & $448\times448$ \\
Visual tokens for the LM & $N/4$ & 1024 & 256 \\
\midrule
LoRA rank & 64 & 64 & 128 \\
LoRA alpha & 128 & 128 & 256 \\
LoRA dropout & 0 & 0 & 0.05 \\
LoRA target modules & LM attn.\ $q,k,v,o$ & LM attn.\ $q,k,v,o$ & All LM linear \\
SFT trainable params (M) & 66 & 24 & 341 \\
AdaLN / AdaLoop params (M) & 82 & 39 & 350 \\
\midrule
Training updates & 40k & 30k & 40k \\
Batch size & \multicolumn{3}{c}{64} \\
Optimizer & \multicolumn{3}{c}{AdamW} \\
Adam $(\beta_1,\beta_2)$ & $(0.9,0.99)$ & $(0.9,0.999)$ & $(0.9,0.95)$ \\
Adam $\epsilon$ & \multicolumn{3}{c}{$10^{-8}$} \\
Learning rate: LoRA & \multicolumn{3}{c}{$1\times10^{-4}$} \\
Learning rate: connector & $5\times10^{-5}$ & $3\times10^{-5}$ & $5\times10^{-5}$ \\
Learning rate: conditioning & \multicolumn{3}{c}{$1\times10^{-4}$} \\
Learning-rate schedule & Cosine to $0.1\times$ & Cosine to $0.1\times$ & Cosine to 0 \\
Warmup updates & 200 & 200 & 1{,}200 \\
Weight decay & $10^{-4}$ & $10^{-4}$ & 0 \\
Gradient clipping norm & \multicolumn{3}{c}{1.0} \\
\midrule
Default loop span & $[2,14)$ & $[3,15)$ & $[2,8)$ \\
Default interface & \multicolumn{3}{c}{Variant A} \\
Training depth $T_{\mathrm{train}}$ & \multicolumn{3}{c}{3} \\
Depth-weight ratio $r$ & 0.6 & 0.8 & 0.6 \\
Depth supervision & \multicolumn{3}{c}{All $T_{\mathrm{train}}$ depths, static geometric weights} \\
Modulation initialization & \multicolumn{3}{c}{Zero output projection} \\
\bottomrule
\end{tabularx}
\endgroup
\end{table}

\paragraph{Backbones and loop spans.} We use Qwen2.5-VL-3B-Instruct, PaliGemma~2~3B~Mix-448, and our re-trained LLaVA-1.5-7B checkpoint at $448\times448$ (Table~\ref{tab:backbone-config}). The notation $[a,b)$ denotes encoder blocks $a,\ldots,b-1$. Each backbone retains its vision--language connector after the vision encoder. Qwen's spatial patch merger reduces each visual state to one quarter as many language-model input tokens, without changing the resolution of the recurrent state. LLaVA applies $2\times2$ average pooling to its 1,024 CLIP patch tokens, producing 256 tokens for the language model.

\begin{figure}[t]
  \centering
  \includegraphics[width=0.9\linewidth]{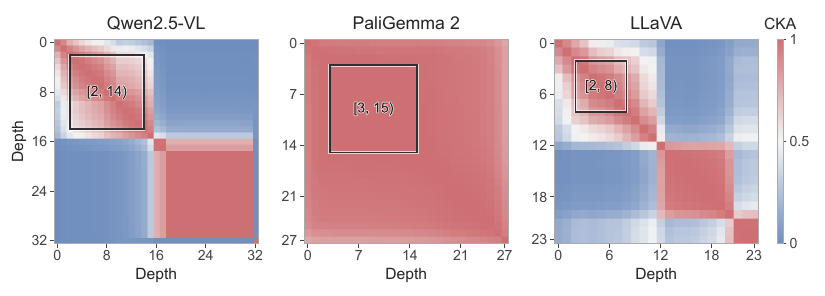}
  \caption{\textbf{Pretrained visual representation similarity.} Linear CKA across depths in Qwen2.5-VL, PaliGemma~2, and LLaVA, using the same $0$--$1$ color scale. Boxes mark the default loop spans in Table~\ref{tab:backbone-config}; their corners correspond to each span's input and output depths. Depth $k$ denotes the representation after $k$ blocks. LLaVA is shown through the penultimate block used by the model. The Qwen matrix is the same as in Figure~\ref{fig:cka-heatmaps}.}
  \label{fig:appendix-cka}
\end{figure}

\Needspace{6\baselineskip}
\paragraph{CKA measurement.}
\label{sec:cka}
Figure~\ref{fig:appendix-cka} shows linear CKA~\citep{kornblith2019similarity} between pretrained vision-encoder representations in Qwen2.5-VL, PaliGemma~2, and LLaVA. We sample 10 images per VisRO-Core6 task and 64 patch-token positions per image, giving 3,840 observations per backbone. Token positions are matched across depths, feature columns are centered.

Depth $0$ denotes the input to the first vision block, and depth $k$ the representation after $k$ blocks. For Qwen, we extract the window-permuted patch embeddings and all 32 block outputs before the visual merger. For PaliGemma, we extract the patch embeddings with positional embeddings and all 27 block outputs before the connector. For LLaVA, we exclude the class token and extract representations after CLIP's pre-layer normalization and through the penultimate block, giving depths $0$--$23$. These representations precede LLaVA's $2\times2$ pooling and projector. CKA characterizes pretrained representation similarity; loop spans are evaluated empirically in Section~\ref{sec:recurrence-ablation}.

\paragraph{Trainable components.} We train the question pooler, block-specific modulation generators, vision--language connector, and language-model LoRA adapters, as well as the iteration embedding of LLaVA AdaLoop described below. The pretrained vision weights remain frozen unless stated otherwise. Variants A--C use the same parameterization. The loop body and modulation generators are shared across iterations, so increasing loop depth adds no parameters.

\paragraph{Question conditioning.} The pooler computes $c_q$ from the question's input embeddings and reuses it across visual iterations. For all three backbones, the question is tokenized separately from the language prompt, without prompt-template, chat-role, or EOS tokens, and truncated to 128 tokens; Qwen and PaliGemma omit BOS, whereas LLaVA prepends it within this budget. The embeddings are detached, converted to FP32, RMS-normalized, and projected from the language-model embedding width (2048, 2304, and 4096 for Qwen, PaliGemma, and LLaVA) to 512. A learned query attends to this sequence through eight-head cross-attention with a residual connection, followed by a residual MLP with RMS normalization and hidden width 1024. Padding tokens are masked, and the pooler is shared across conditioned blocks. Images, answer targets, and generated drafts do not enter the question condition.

\begin{figure}[t]
  \centering
  \begin{minipage}[c]{0.41\linewidth}
    \centering
    \includegraphics[width=\linewidth]{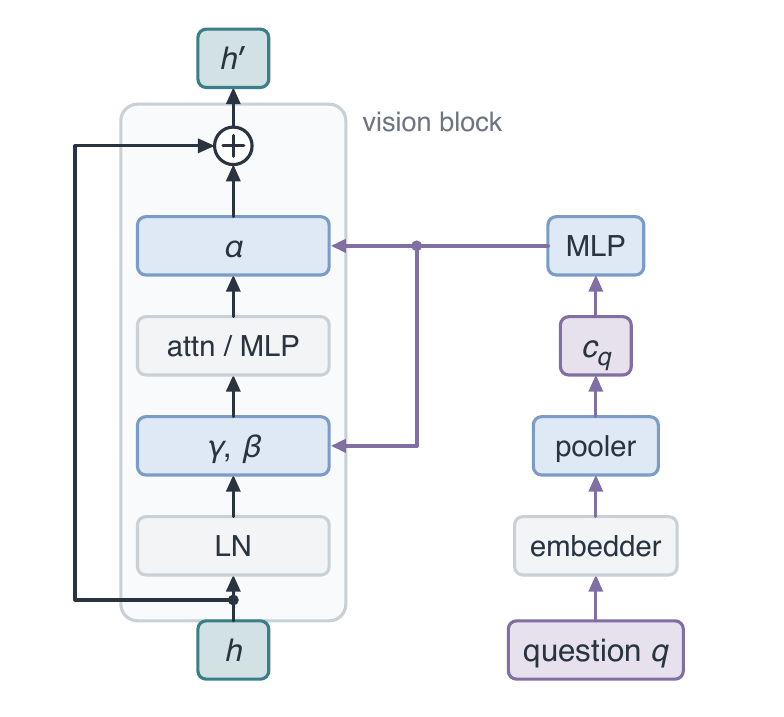}
  \end{minipage}\hfill
  \begin{minipage}[c]{0.50\linewidth}
    \caption{\textbf{Question-conditioned visual updates.}
    One residual branch of a frozen vision block (gray).
    The question is pooled into $c_q$ and mapped to branch-specific
    shifts, scales, and gains. Coefficients are shared across visual
    tokens and, except in LLaVA AdaLoop, across iterations; the multiplicative factors are
    $\mathbf{1}+\gamma$ and $\mathbf{1}+\alpha$
    (Equation~\ref{eq:adaln}). Blue modules are trainable. Only each
    generator's output layer is zero-initialized, preserving the
    pretrained computation at initialization.}
    \label{fig:adaloop-conditioning}
  \end{minipage}
\end{figure}

\paragraph{Modulation and initialization.} Each block has its own generator $A_\ell$, comprising layer normalization and a two-layer MLP with a SiLU activation. The output supplies a shift, scale, and residual gain for each of the attention and MLP branches, as in Equation~\ref{eq:adaln}. Coefficients are broadcast over visual tokens. For Qwen and PaliGemma, they are also shared across visual iterations, as stated in Section~\ref{sec:vision-loop}. LLaVA AdaLoop is an exception: $c_q$ is still computed once per example, but a learned embedding of the iteration index is added to it before each iteration, so the coefficients vary across iterations. The embedding maps sine and cosine features of $t-1$, at 32 geometrically spaced frequencies from $1$ to $1/32$, through a two-layer MLP with a SiLU activation and width 512 (0.3M parameters). LLaVA single-pass AdaLN has no such embedding. We zero-initialize only the generator's output layer, so the first visual iteration initially recovers the pretrained encoder computation. The multiplicative factors are not squashed or clipped. Figure~\ref{fig:adaloop-conditioning} illustrates one modulated branch.

\subsection{Training}
\label{app:training-settings}

\begin{figure}[t]
  \centering
  \includegraphics[width=0.93\linewidth]{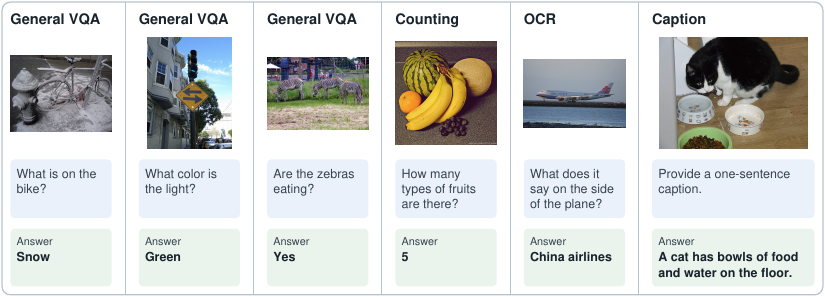}
  \caption{\textbf{Examples from the fixed LLaVA-NeXT training subset.}
  The examples illustrate general visual question answering, counting,
  optical character recognition, and captioning. Answers use the original
  dataset annotations~\citep{liu2024llavanext}; prompts are shortened
  for readability.}
  \label{fig:llava-next-examples}
\end{figure}

\paragraph{Data.} The six tasks used in our experiments each contribute 100k training examples, 600 ID test examples, and 600 OOD test examples. Training and ID tests contain 25k and 150 examples per task and difficulty level, respectively. All task images are rendered at $448\times448$ before model-specific preprocessing. For Count Clusters, both test splits use icon classes held out from training; ID examples retain the training difficulty range. We mix the 600k task examples with a fixed 100k-example subset of LLaVA-NeXT, using its original answer annotations (Figure~\ref{fig:llava-next-examples}).

\paragraph{Optimization.} Table~\ref{tab:backbone-config} specifies the optimization settings, with learning rates listed separately for the language-model LoRA adapters, vision--language connector, and question-conditioning modules. The latter include the question pooler, the modulation generators, and, for LLaVA AdaLoop, the iteration embedding. Within each backbone, SFT, single-pass AdaLN, and AdaLoop use the same training mixture and optimizer-update budget. Corresponding adapter initializations and training-example order are paired by seed.

\paragraph{Depth supervision.} At each supervised depth, we teacher-force the same final answer after the variant-specific prefix $\Pi_t^{(v)}$. Only target-answer tokens contribute to the loss; question, visual, separator, and draft tokens are excluded. We use normalized depth weights $w_t\propto r^{T_{\mathrm{train}}-t}$, where $0<r<1$, so later depths receive more weight. We use static geometric weights with $r=0.6$ for Qwen and LLaVA and $r=0.8$ for PaliGemma. Deep supervision evaluates the weighted losses at all trained depths. In our Qwen depth-sampling experiments, we instead sample a depth with probability $w_t$ and compute its unweighted mean answer-token loss. This gives an unbiased estimate of the same objective in Section~\ref{sec:loss}.

\paragraph{Drafts and gradient flow.} Variant C generates its drafts greedily from the model's predictions. Drafts enter subsequent language prefixes without being replaced by the target answer or differentiated through token selection. We perform a fresh language-model prefill between rounds and detach the question embeddings used for conditioning. In all variants, gradients from the supervised loss pass through frozen vision blocks to the conditioning modules. Appendix~\ref{app:abc-cost} accounts for the additional draft computation.

\subsection{Inference}
\label{app:inference-settings}

\paragraph{Inputs and decoding.} On VisRO, we supply the task question through the language interfaces in Section~\ref{sec:interfaces}. The visual question conditioner receives only the original question text, excluding evaluation-only answer-format instructions. We use backbone-specific causal/prefix attention conventions with the masking required by each readout variant. Final answers are decoded greedily until EOS or a maximum of 64 newly generated tokens for Qwen, 128 for PaliGemma and 32 for LLaVA.

\paragraph{Answer-stability stopping.} Table~\ref{tab:main-results} uses $p=1$ and maximum depth $T_{\mathrm{test}}=3$. We obtain predictions from separate fixed-depth evaluations at depths one through three, then apply the stopping rule offline. After lowercasing answers and normalizing whitespace, we select the depth-two answer if it agrees with the depth-one answer, and otherwise select the depth-three answer. This rule uses no target answers.

\paragraph{Evaluation depth.} Figure~\ref{fig:abc-loop-depth} evaluates fixed depths one through six without stopping, using models trained with $T_{\mathrm{train}}=3$. The mean depth $\bar{T}$ in Table~\ref{tab:main-results} is the mean stopping depth selected by the rule. Appendix~\ref{app:abc-cost} estimates the cost of executing this rule online, including intermediate answer generation.

\Needspace{23\baselineskip}
\subsection{Baseline Implementations}
\label{app:baseline-implementations}

\paragraph{Single-pass controls.} Zero-shot evaluation uses the pretrained VLM. SFT trains the vision--language connector and language-model LoRA adapters while keeping the vision encoder frozen. Single-pass AdaLN adds AdaLoop's question pooler and modulation generators and applies the same visual span once. Its trainable parameter count matches AdaLoop's within each backbone, except for the 0.3M-parameter iteration embedding of LLaVA AdaLoop. The question-independent control in Table~\ref{tab:component-ablations} supplies the same null question for all training and test examples while retaining the modulation modules.

\paragraph{LIVR.} Our Qwen implementation of LIVR~\citep{li2026latent} matches SFT in initialization, the 700k-example training mixture, sampling seed, global batch size, rank-64 language LoRA, trainable vision merger, and total 40k-update budget. Following the paper's best ablation, we use $K=16$ learned latent tokens and train for 16k updates with visual-bottleneck attention, followed by 24k updates with standard attention. Training uses only final-answer supervision.

\paragraph{SFT+GRPO.} We implement GRPO~\citep{shao2024deepseekmath} using EasyR1~\citep{zheng2025easyr1}, initialized from the converged Qwen SFT checkpoint and rewarded by exact final-answer accuracy. This warm start follows preliminary runs from the pretrained checkpoint, which produced almost no correct rollouts and consequently little within-group reward variation. The GRPO stage adds training beyond the SFT budget.

\FloatBarrier

\section{Additional Experiments and Analyses}
\label{app:additional-experiments}

\begin{table}[!t]
\centering
\caption{\textbf{Fixed-depth evaluation on VisRO and general VQA.}
Scores are accuracies (\%); VisRO scores are macro averages across six
tasks. Params denotes trainable parameters (M).
$^{*}$General VQA evaluation uses fixed $T=1$.}
\label{tab:fixed-depth-results}
\begingroup
\fontsize{8}{10}\selectfont
\setlength{\tabcolsep}{6pt}
\renewcommand{\arraystretch}{1.13}
\newlength{\fixeddepthoverallwidth}
\settowidth{\fixeddepthoverallwidth}{\textbf{Overall}}
\begin{tabular}{lcc*{2}{>{\columncolor{black!5}\centering\arraybackslash}m{\fixeddepthoverallwidth}}cc}
\toprule
\multirow{2}{*}{Method} & \multirow{2}{*}{Params}
  & \multicolumn{3}{c}{VisRO overall}
  & \multicolumn{2}{c}{General VQA$^{*}$} \\
\cmidrule(l{2pt}r{2pt}){3-5}\cmidrule(l{2pt}r{2pt}){6-7}
  & & $T$ & \textbf{ID} & \textbf{OOD} & MMStar & MMBench \\
\midrule
\multicolumn{7}{@{}l}{\textit{Qwen2.5-VL-3B}} \\
\addlinespace[2pt]
SFT & 66 & 1 & 49.9 & 23.9 & 51.7 & 77.1 \\
Single-pass AdaLN & 82 & 1 & 60.0 & 30.2 & 51.1 & 75.4 \\
\textbf{AdaLoop-A} & 82 & 3 & \textbf{63.7} & \textbf{32.4} & 51.1 & 75.5 \\
\midrule
\multicolumn{7}{@{}l}{\textit{PaliGemma2-3B}} \\
\addlinespace[2pt]
SFT & 24 & 1 & 62.6 & 31.5 & 46.1 & 72.0 \\
Single-pass AdaLN & 39 & 1 & 72.4 & 40.0 & 44.7 & 68.6 \\
\textbf{AdaLoop-A} & 39 & 3 & \textbf{74.4} & \textbf{44.0} & 44.3 & 71.2 \\
\midrule
\multicolumn{7}{@{}l}{\textit{LLaVA-1.5-7B}} \\
\addlinespace[2pt]
SFT & 341 & 1 & 50.4 & 24.8 & 33.7 & 59.1 \\
Single-pass AdaLN & 350 & 1 & 57.7 & 29.6 & 34.8 & 60.8 \\
\textbf{AdaLoop-A} & 350 & 3 & \textbf{62.1} & \textbf{31.0} & 33.7 & 60.1 \\
\bottomrule
\end{tabular}
\endgroup
\end{table}

\begin{table}[!t]
\centering
\caption{\textbf{Operation counts for online inference.}
All variants execute the visual loop $t$ times at fixed depth
and $\tau$ times under answer-stability stopping.
A readout applies the coda and connector to a visual state.
A generation includes language-model prefill and answer decoding.
Adaptive stopping checks the answer at every executed depth.}
\label{tab:abc-cost}
\begingroup
\fontsize{8}{10}\selectfont
\setlength{\tabcolsep}{6pt}
\renewcommand{\arraystretch}{1.1}
\begin{tabular}{lcccc}
\toprule
\multirow{2}{*}{Variant}
  & \multicolumn{2}{c}{Fixed depth $t$}
  & \multicolumn{2}{c}{Adaptive depth $\tau$} \\
\cmidrule(lr){2-3}\cmidrule(lr){4-5}
  & Readouts & Generations & Readouts & Generations \\
\midrule
A & 1   & 1   & $\tau$ & $\tau$ \\
B & $t$ & 1   & $\tau$ & $\tau$ \\
C & $t$ & $t$ & $\tau$ & $\tau$ \\
\bottomrule
\end{tabular}
\endgroup
\end{table}

We supplement the main experiments with fixed-depth evaluation,
training on all 24 VisRO tasks, and computational cost.

\subsection{Fixed-Depth Evaluation}
\label{app:fixed-depth-results}

Table~\ref{tab:fixed-depth-results} reports fixed-depth results
for the same trained checkpoints evaluated in
Table~\ref{tab:main-results}.
AdaLoop uses $T=3$ on VisRO, while all general VQA evaluations
use $T=1$.
At fixed depth, AdaLoop improves over single-pass AdaLN on
both VisRO-ID and VisRO-OOD across all three backbones.
Changes on general VQA are mixed across models and benchmarks.

\subsection{Training on All 24 VisRO Tasks}
\label{app:visro24-training}

\begin{table}[!t]
\centering
\caption{\textbf{Training on all 24 VisRO tasks.}
Models are trained on all 24 tasks for 160k (Qwen2.5-VL, LLaVA) or 120k (PaliGemma~2) updates and evaluated on the full 24-task ID and OOD test sets (600 examples per task and split).
ID columns report macro-average accuracy (\%) over the tasks in each category of Table~\ref{tab:benchmark-tasks} (\emph{Disc.}: discrimination and grouping; \emph{Paths}: paths and connectivity; \emph{Planar}: planar transformations and composition; \emph{Spatial}: spatial transformations and structure); \emph{Overall} is the macro average over all 24 tasks.
AdaLoop uses the same variant and stopping rule as Table~\ref{tab:main-results}; $\bar{T}$: mean selected stopping depth on the ID set.
Per-task ID results are in Table~\ref{tab:visro24-training-id}.}
\label{tab:visro24-training}
\begingroup
\fontsize{8}{10}\selectfont
\setlength{\tabcolsep}{5pt}
\renewcommand{\arraystretch}{1.1}
\newlength{\visrotrainoverallwidth}
\settowidth{\visrotrainoverallwidth}{\textbf{Overall}}
\newlength{\visrotraincatwidth}
\settowidth{\visrotraincatwidth}{Spatial}
\begin{tabular}{lc*{4}{>{\centering\arraybackslash}p{\visrotraincatwidth}}*{2}{>{\columncolor{black!5}\centering\arraybackslash}p{\visrotrainoverallwidth}}}
\toprule
\multirow{2}{*}{Method} & \multirow{2}{*}{$\bar{T}$} & \multicolumn{5}{c}{ID} & \multicolumn{1}{c}{OOD} \\
\cmidrule(l{2pt}r{2pt}){3-7}\cmidrule(l{2pt}r{2pt}){8-8}
 & & Disc. & Paths & Planar & Spatial & \textbf{Overall} & \textbf{Overall} \\
\midrule
\multicolumn{8}{@{}l}{\textit{Qwen2.5-VL-3B}} \\
\addlinespace[2pt]
SFT & 1.0 & 59.8 & 47.0 & 41.1 & 75.6 & 57.6 & 17.9 \\
Single-pass AdaLN & 1.0 & 73.7 & 53.0 & 48.5 & 84.4 & 67.1 & 23.0 \\
\textbf{AdaLoop} & 2.2 & 75.4 & 57.8 & 51.3 & 85.8 & \textbf{69.5} & \textbf{25.2} \\
\midrule
\multicolumn{8}{@{}l}{\textit{PaliGemma2-3B}} \\
\addlinespace[2pt]
SFT & 1.0 & 73.7 & 53.1 & 47.7 & 68.6 & 62.3 & 18.8 \\
Single-pass AdaLN & 1.0 & 78.2 & 64.0 & 50.9 & 70.3 & 66.7 & 23.8 \\
\textbf{AdaLoop} & 2.1 & 86.9 & 66.7 & 51.4 & 70.5 & \textbf{69.9} & \textbf{24.5} \\
\midrule
\multicolumn{8}{@{}l}{\textit{LLaVA-1.5-7B}} \\
\addlinespace[2pt]
SFT & 1.0 & 63.9 & 48.0 & 31.7 & 60.6 & 52.2 & 17.3 \\
Single-pass AdaLN & 1.0 & 74.3 & 52.2 & 33.6 & 68.4 & 58.7 & 19.1 \\
\textbf{AdaLoop} & 2.1 & 76.9 & 53.5 & 36.1 & 69.4 & \textbf{60.6} & \textbf{21.1} \\
\bottomrule
\end{tabular}
\endgroup
\end{table}

\begin{table}[!t]
\centering
\caption{\textbf{Per-task ID accuracy (\%) after training on all 24 VisRO tasks.}
Same models as Table~\ref{tab:visro24-training}; \emph{AdaLN} denotes single-pass AdaLN. Bold: best overall accuracy for each backbone.}
\label{tab:visro24-training-id}
\begingroup
\fontsize{8}{10}\selectfont
\setlength{\tabcolsep}{2pt}
\renewcommand{\arraystretch}{1.02}
\newlength{\visrotrainidcolwidth}
\settowidth{\visrotrainidcolwidth}{AdaLoop}
\begin{tabular*}{\linewidth}{@{\extracolsep{\fill}}l*{9}{>{\centering\arraybackslash}p{\visrotrainidcolwidth}}@{}}
\toprule
\multirow{2}{*}{Task} & \multicolumn{3}{c}{Qwen2.5-VL-3B} & \multicolumn{3}{c}{PaliGemma2-3B} & \multicolumn{3}{c}{LLaVA-1.5-7B} \\
\cmidrule(l{2pt}r{2pt}){2-4}\cmidrule(l{2pt}r{2pt}){5-7}\cmidrule(l{2pt}r{2pt}){8-10}
 & SFT & AdaLN & AdaLoop & SFT & AdaLN & AdaLoop & SFT & AdaLN & AdaLoop \\
\midrule
\multicolumn{10}{@{}l}{\emph{Discrimination and grouping}} \\
Count clusters & 58.5 & 65.3 & 67.5 & 72.3 & 82.7 & 85.3 & 48.0 & 53.2 & 56.8 \\
Points on lines & 44.7 & 50.0 & 51.8 & 46.2 & 70.8 & 69.0 & 51.0 & 68.2 & 75.8 \\
Find the same & 69.5 & 91.7 & 95.3 & 87.7 & 96.8 & 98.2 & 85.0 & 93.3 & 96.5 \\
Find the different & 80.8 & 98.8 & 99.3 & 98.0 & 100.0 & 99.8 & 87.2 & 98.7 & 99.7 \\
Find the shadow & 81.7 & 97.8 & 98.7 & 95.5 & 54.2 & 100.0 & 89.2 & 99.7 & 99.5 \\
Overlapping glyphs & 56.5 & 86.5 & 88.8 & 84.5 & 93.3 & 93.3 & 64.2 & 83.8 & 86.7 \\
Overlap matching & 26.7 & 25.8 & 26.5 & 31.7 & 49.5 & 62.5 & 22.8 & 23.3 & 23.5 \\
\midrule
\multicolumn{10}{@{}l}{\emph{Paths and connectivity}} \\
Flow Free & 12.5 & 22.2 & 31.2 & 24.3 & 35.2 & 42.7 & 17.5 & 31.0 & 27.2 \\
Metro map & 57.7 & 61.5 & 64.2 & 64.0 & 71.0 & 73.3 & 49.2 & 51.8 & 53.2 \\
Maze & 58.8 & 63.5 & 65.3 & 68.5 & 88.7 & 92.7 & 60.5 & 61.3 & 65.0 \\
Path tracing & 59.2 & 64.8 & 70.5 & 55.7 & 61.2 & 58.0 & 64.7 & 64.5 & 68.5 \\
\midrule
\multicolumn{10}{@{}l}{\emph{Planar transformations and composition}} \\
Jigsaw completion & 56.3 & 85.5 & 92.0 & 84.8 & 92.7 & 91.7 & 0.2 & 0.5 & 0.3 \\
Reconstruction & 26.0 & 23.3 & 26.0 & 24.7 & 25.3 & 25.0 & 24.7 & 25.5 & 26.0 \\
Tangram reconstruction & 44.2 & 58.3 & 62.5 & 51.0 & 62.5 & 65.5 & 59.7 & 66.7 & 69.7 \\
Mirroring patterns & 75.0 & 75.0 & 77.3 & 74.7 & 74.7 & 75.0 & 75.0 & 75.0 & 75.0 \\
Rotation patterns & 0.8 & 1.0 & 1.0 & 1.8 & 0.3 & 1.5 & 0.8 & 0.8 & 1.5 \\
Logic matrix & 44.2 & 47.7 & 48.8 & 49.0 & 49.7 & 50.0 & 29.8 & 32.8 & 43.8 \\
\midrule
\multicolumn{10}{@{}l}{\emph{Spatial transformations and structure}} \\
3D block counting & 59.5 & 89.0 & 90.7 & 84.0 & 90.8 & 90.5 & 76.5 & 91.8 & 91.8 \\
Block changes & 74.0 & 94.7 & 97.5 & 98.5 & 99.7 & 100.0 & 87.7 & 95.2 & 97.8 \\
Cube unfolding & 25.2 & 13.2 & 16.3 & 7.7 & 4.3 & 4.2 & 2.7 & 2.5 & 2.2 \\
Paper folding & 100.0 & 99.8 & 100.0 & 1.2 & 2.0 & 2.0 & 1.5 & 1.0 & 1.3 \\
3D views & 99.0 & 100.0 & 99.7 & 99.0 & 99.0 & 99.5 & 96.5 & 99.2 & 99.7 \\
View reconstruction & 96.7 & 99.7 & 99.8 & 99.0 & 99.5 & 99.7 & 91.8 & 99.3 & 99.5 \\
Viewpoint rotation & 75.0 & 94.7 & 96.8 & 91.0 & 97.0 & 97.7 & 67.5 & 89.8 & 93.7 \\
\midrule
\textbf{Overall} & 57.6 & 67.1 & \textbf{69.5} & 62.3 & 66.7 & \textbf{69.9} & 52.2 & 58.7 & \textbf{60.6} \\
\bottomrule
\end{tabular*}
\endgroup
\end{table}

Table~\ref{tab:visro24-training} trains the same three methods on all
24 VisRO tasks instead of Core6, and
Table~\ref{tab:visro24-training-id} reports per-task ID accuracy.
As on Core6, single-pass AdaLN improves over SFT on both ID and OOD for
all three backbones, and AdaLoop further improves over AdaLN on both
ID and OOD for all three backbones.

\subsection{Computational Cost}
\label{app:abc-cost}
We describe the FLOP accounting for the Qwen2.5-VL
Variant-A comparisons in Section~\ref{sec:compute-limitations}.

\paragraph{Counting convention.} We count one multiply--add as two FLOPs and include matrix multiplications in the vision encoder, conditioning modules, connector, and language model. Each answer uses a KV cache, with a fresh prefill at each depth. We exclude padding, elementwise operations, normalization, softmax, activation-checkpoint recomputation, optimizer updates, data processing, and communication. These are arithmetic estimates, not latency measurements.

\begin{table}[!t]
\centering
\captionsetup{position=bottom}

\setlength{\tabcolsep}{6pt}
\renewcommand{\arraystretch}{1.1}
\captionsetup[subtable]{position=bottom,font=footnotesize,labelfont=normalfont,skip=4pt}
\begin{subtable}{\linewidth}
\centering
\fontsize{8}{10}\selectfont
\begin{tabular}{@{}lccc@{}}
\toprule
Method & Depth & TFLOPs/example & Relative FLOPs \\
\midrule
Single-pass AdaLN & 1 & 3.133 & $1.00\times$ \\
AdaLoop (fixed) & 3 & 4.170 & $1.33\times$ \\
AdaLoop (adaptive) & $\leq 3$ & 6.777 & $2.16\times$ \\
\bottomrule
\end{tabular}
\vspace{5pt}
\caption{\textbf{Inference on VisRO-ID.}
Costs average 3,600 examples using text-derived generation lengths.
Adaptive costs assume online stopping with $p=1$; ratios are relative to single-pass AdaLN.}
\label{tab:measured-flops-inference}
\end{subtable}

\medskip
\begin{subtable}{\linewidth}
\centering
\fontsize{8}{10}\selectfont
\begin{tabular}{@{}lrccc@{}}
\toprule
Method & Updates & TFLOPs/example & Total EFLOPs & Relative FLOPs \\
\midrule
Single-pass AdaLN & 40k & 6.985 & 17.88 & $1.00\times$ \\
\midrule
Single-pass AdaLN & 50k & 6.985 & 22.35 & $1.25\times$ \\
AdaLoop (sampled) & 40k & 9.229 & 23.63 & $1.32\times$ \\
\midrule
Single-pass AdaLN & 120k & 6.985 & 53.64 & $3.00\times$ \\
AdaLoop (deep sup.) & 40k & 19.981 & 51.15 & $2.86\times$ \\
\bottomrule
\end{tabular}
\vspace{5pt}
\caption{\textbf{Training on the 700k-example mixture.}
Costs include forward and backward computation with global batch size 64.
The sampled estimate assumes per-example state masking.
Total-cost ratios are relative to single-pass AdaLN at 40k updates.}
\label{tab:measured-flops-training}
\end{subtable}
\caption{\textbf{Estimated computational cost on Qwen2.5-VL.}
AdaLoop uses variant A with loop span $[2,14)$.
Costs exclude padding and activation-checkpoint recomputation;
TFLOPs and EFLOPs denote $10^{12}$ and $10^{18}$ operations.}
\label{tab:measured-flops}
\end{table}

\paragraph{Inference cost.} Table~\ref{tab:abc-cost} counts operations for fixed-depth and online adaptive inference. Adaptive costs include one visual trajectory and answer generation at each depth through the selected stopping depth. They exclude repeated fixed-depth passes used for offline evaluation (Appendix~\ref{app:inference-settings}). Table~\ref{tab:measured-flops-inference} averages per-example estimates over 3,600 VisRO-ID examples, accounting for each example's context and generation lengths. We estimate generated lengths by re-tokenizing decoded predictions and adding EOS, capped at 64 tokens; these may differ from the original token counts. Online stopping uses $p=1$ and a depth limit of three, with mean selected depth 2.19.

\paragraph{Training cost.}
Deep supervision shares one visual computation graph across
depths, while each supervised depth incurs its own readout
and language-model pass.
The cost estimate includes backward computation through
frozen vision blocks wherever it is needed to train the
conditioning modules.
For matrix multiplications, forward-only computation is counted
once; forward plus an input gradient is counted twice; and forward
plus gradients for both operands is counted three times.
Thus, frozen linear layers that propagate input gradients cost
twice their forward computation, while trainable linear layers
and attention matrix products requiring gradients for both
operands cost three times.
LoRA computation is included during training and merged for inference.
Intermediate language-model output heads are applied only at
supervised token positions; the final head is applied to the full
sequence, following the implementation.

For depth-sampled supervision, we estimate the cost of
the example-masked implementation.
It masks state updates after each example's sampled depth
but still executes all iterations up to the training limit.
The estimate includes this visual computation and one
supervised readout at the sampled depth.

We estimate total training FLOPs by averaging per-example costs over
the training mixture and multiplying by the global batch size and
number of updates (Table~\ref{tab:measured-flops-training}).
The sampled and deeply supervised runs use 5.7\% more and 4.6\% less
compute than their respective AdaLN controls.

\begin{table}[!t]
\centering
\captionsetup{position=bottom,skip=6pt}
\captionsetup[subtable]{position=bottom,font=footnotesize,labelfont=normalfont,skip=3pt}
\begin{subtable}{\linewidth}
\centering
\begingroup
\fontsize{8.5}{10.5}\selectfont
\setlength{\tabcolsep}{4pt}
\renewcommand{\arraystretch}{1.1}
\begin{tabularx}{\linewidth}{ll>{\raggedright\arraybackslash}X}
\toprule
Method & Repeated component & Initialization / adaptation \\
\midrule
\multicolumn{3}{l}{\emph{Vision models}} \\
\addlinespace[2pt]
SR-ViT~\citep{lee2026soft}
  & ViT blocks
  & From scratch on ImageNet \\
\addlinespace[3pt]
LoopViT~\citep{shu2026loopvit}
  & Hybrid ViT blocks
  & Visual ARC pretraining; test-time training \\
\addlinespace[3pt]
Raptor~\citep{jacobs2026block}
  & Surrogate ViT blocks
  & Fit pretrained ViT trajectories \\
\midrule
\multicolumn{3}{l}{\emph{Language models}} \\
\addlinespace[2pt]
Huginn~\citep{geiping2026scaling}
  & Middle LM blocks
  & From scratch \\
\addlinespace[3pt]
Ouro~\citep{zhu2025ouro}
  & Full LM blocks
  & From scratch \\
\addlinespace[3pt]
LoopUS~\citep{park2026loopus}
  & Middle LM blocks
  & Continued LM pretraining \\
\midrule
\multicolumn{3}{l}{\emph{Multimodal models}} \\
\addlinespace[2pt]
RecursiveVLM~\citep{xu2026looping}
  & Full LM blocks
  & Continued VLM pretraining + SFT \\
\addlinespace[3pt]
HIVE~\citep{hive2026}
  & Huginn recurrent core
  & Pretrained vision + LM; multimodal training \\
\addlinespace[3pt]
CVRR~\citep{park2026cvrr}
  & Single LM block
  & Pretrained VLM; recurrent LoRA \\
\addlinespace[3pt]
\rowcolor{black!5}
\textbf{AdaLoop}
  & Frozen vision blocks
  & Train conditioning, connector, LM LoRA \\
\bottomrule
\end{tabularx}
\endgroup
\caption{\textbf{Looped architectures.}
Parameters are shared across repeated applications of the listed component.}
\label{tab:looped-architectures}
\end{subtable}

\medskip
\begin{subtable}{\linewidth}
\centering
\begingroup
\fontsize{8.5}{10.5}\selectfont
\setlength{\tabcolsep}{4pt}
\renewcommand{\arraystretch}{1.1}
\begin{tabularx}{\linewidth}{ll>{\raggedright\arraybackslash}X}
\toprule
Method & Mechanism & Conditioning / feedback \\
\midrule
Mema~\citep{liu2026mema}
  & Cross-layer memory + modulation
  & Input question \\
\addlinespace[3pt]
VPT~\citep{yu2025vpt}
  & Re-encoding; conditioned projection
  & LM control-token state \\
\addlinespace[3pt]
\citet{galatolo2026lightweight}
  & Feedback to encoder inputs
  & LM image-token states \\
\addlinespace[3pt]
DVR~\citep{zhang2026dvr}
  & Re-encoding
  & Generated sub-questions \\
\addlinespace[3pt]
\rowcolor{black!5}
\textbf{AdaLoop}
  & Recurrent visual updates
  & Input question \\
\bottomrule
\end{tabularx}
\endgroup
\caption{\textbf{Language-guided visual processing.}
Visual computation can be conditioned directly on the question or on
intermediate language-model states or outputs.}
\label{tab:visual-feedback}
\end{subtable}

\caption{\textbf{Looped architectures and language-guided visual processing.}
The two panels compare complementary architectural choices across different
tasks and training settings. LM denotes language model.}
\label{tab:looped_comparison}
\end{table}

\section{Additional Related Work}
\label{app:looped-comparison}

\paragraph{Reusing pretrained visual computation.}
Table~\ref{tab:looped-architectures} compares recurrent architectures,
and Table~\ref{tab:visual-feedback} compares how language guides
visual processing.
Raptor~\citep{jacobs2026block} learns recurrent surrogate blocks
to approximate pretrained ViT activation trajectories.
Mema~\citep{liu2026mema} maintains question-conditioned memory
across encoder layers.
AdaLoop repeatedly applies the original frozen vision blocks
to an evolving visual state.
Question-conditioned modulation guides these updates, with generator
parameters shared across iterations.

\paragraph{Relation to procedural training resources.}
VisRO shares with VBVR and VBVR-Pro an emphasis on procedural
data generation and programmatic
evaluation~\citep{wang2026vbvr,xu2026vbvrpro}.
These works evaluate generated visual solutions.
Our experiments use image--question--answer triples and
final-answer supervision, without prescribed visual solutions
or intermediate trajectories.

\section{VisRO: Tasks, Generation, and Evaluation}
\label{app:dataset}

VisRO is a procedural dataset and generation pipeline with 24 visual reasoning tasks spanning counting, discrimination, connectivity, geometric transformations, and spatial reasoning. This appendix describes the tasks, their difficulty controls, and the data generation and evaluation protocol. The dataset is available at \url{https://huggingface.co/datasets/Lyy0725/VisRO}, and the generators and scoring code at \url{https://github.com/Lyy-iiis/VisRO}.

\subsection{Task Coverage and Definitions}
\label{app:benchmark-tasks}

Table~\ref{tab:benchmark-tasks} groups the 24 tasks by the visual operations they require: discrimination and grouping, paths and connectivity, planar transformations and composition, and spatial transformations and structure. Figures~\ref{fig:benchmark}, \ref{fig:benchmark-extended}, and~\ref{fig:benchmark-extended-more} together illustrate all tasks. Questions specify both the requested operation and the answer format, which is often structured rather than a choice (e.g., per-segment counts, piece placements, or queried-cell values).

\begin{table}[!t]
\centering
\caption{\textbf{VisRO task coverage.}
The 24 tasks are grouped by the visual operations they require.
Questions are abbreviated examples; several tasks mix question families,
and answer forms are those of the released data.}
\label{tab:benchmark-tasks}
\begingroup
\vspace{4pt}
\small
\setlength{\tabcolsep}{4pt}
\renewcommand{\arraystretch}{1.08}
\begin{tabularx}{\linewidth}{@{}l>{\raggedright\arraybackslash}Xl@{}}
\toprule
Task & Representative question & Answer form \\
\midrule
\multicolumn{3}{@{}l}{\emph{Discrimination and grouping}} \\
Count clusters & How many objects overlap another object? & Integer \\
Points on lines & Count dots per segment and distinct dots. & Integer list \\
Find the same & Which items are identical? & Correspondences \\
Find the different & Which item differs from the others? & Position \\
Find the shadow & Match shadows to objects; list unmatched ones. & Correspondences \\
Overlapping glyphs & Read symbols in the indicated color order. & Symbol sequence \\
Overlap matching & Which outline occurs in the superposition? & Choice \\
\addlinespace[4pt]
\multicolumn{3}{@{}l}{\emph{Paths and connectivity}} \\
Flow Free & Count cells or turns along one or all pipes. & Integer(s) \\
Metro map & Fewest stops or transfers? Nearest ringed station? & Integer or name \\
Maze & Rank or count starts by path distance to a target. & Choice or integer \\
Path tracing & What are the marked starts connected to? & Labels or pairs \\
\addlinespace[4pt]
\multicolumn{3}{@{}l}{\emph{Planar transformations and composition}} \\
Jigsaw completion & Which piece and rotation fill each hole? & Pieces with turns \\
Reconstruction & Which set of pieces can form the target? & Choice \\
Tangram reconstruction & Where does each piece go to cover the shape? & Placements \\
Mirroring patterns & Which colors do the queried cells have? & Color sequence \\
Rotation patterns & Which figure shows each requested rotation? & Indices or `none' \\
Logic matrix & Complete the matrix or apply the composed rules. & Choice or grid \\
\addlinespace[4pt]
\multicolumn{3}{@{}l}{\emph{Spatial transformations and structure}} \\
3D block counting & What minimum cube count fits the view? & Integer \\
Block changes & How many cubes were added or removed? & Integer(s) \\
Cube unfolding & Which cube views fold from this net? & Index set \\
Paper folding & Which queried cells have holes after unfolding? & Bit string \\
3D views & Which projection matches the viewpoint? & Choice \\
View reconstruction & Which structure fits the given views? & Choice \\
Viewpoint rotation & Which candidate is the rotated reference? & Choice \\
\bottomrule
\end{tabularx}
\endgroup
\end{table}

\begin{table}[!t]
\centering
\caption{\textbf{Example ID and OOD settings in VisRO.}
We compare xhard, the highest ID difficulty, with the held-out OOD
settings. A trailing + denotes a lower bound. In cube unfolding, OOD items
distinguish same-color faces by glyph identity and orientation.}
\label{tab:benchmark-difficulty}
\vspace{4pt}
\begingroup
\small
\setlength{\tabcolsep}{4pt}
\renewcommand{\arraystretch}{1.15}
\begin{tabular*}{\linewidth}{@{\extracolsep{\fill}}lll@{}}
\toprule
Task & ID (xhard) & OOD \\
\midrule
Find the same & 20 icons; 6 matching pairs; 6 near matches & 25 icons; 8 matching pairs; 8 near matches \\
Find the different & 9\texttimes{}9 grid; local alteration & 10\texttimes{}10 grid; smaller alteration cap \\
Cube unfolding & 10 views; orientation distractors & 12 views; glyph-orientation cues \\
3D block counting & 4\texttimes{}5 base; max height 5; min count 30+ & 5\texttimes{}5 base; max height 6; min count 45+ \\
Block changes & 8--10 changes; 5+ columns & 12--16 changes; 8+ columns \\
Viewpoint rotation & 4\texttimes{}5 base; 20--28 cubes & 5\texttimes{}5 base; 30--42 cubes \\
\bottomrule
\end{tabular*}
\endgroup
\end{table}

\subsection{Difficulty Levels and OOD Evaluation}
\label{app:benchmark-difficulty}

\paragraph{Difficulty levels.} Training and in-distribution (ID) evaluation use four task-specific difficulty levels: easy, medium, hard, and xhard. ID evaluation uses independently generated test examples from these four levels. Difficulty is controlled through generation parameters such as object count, overlap, path length, crossings, and distractor similarity. Table~\ref{tab:benchmark-difficulty} compares xhard, the highest ID level, with the OOD settings for representative tasks. Tier names describe task-specific settings and do not imply equal difficulty across tasks.

Difficulty controls also account for the rendered problem. For example, near-matching icons increase comparison demands, while cube-net distractors can differ in face orientation. These changes must preserve the visual cues needed to determine the answer. Rendering resolutions are task-specific and specified by the generator configurations.

\paragraph{The OOD split.} Out-of-distribution (OOD) examples introduce task-specific increases in scene complexity or shifts in generation rules and visual appearance. For example, 3D block counting increases the base dimensions, maximum height, and minimum cube count (Table~\ref{tab:benchmark-difficulty}). Flow Free holds out a path-construction program and rendering or material combinations; Metro Map uses four held-out rendering styles. Both tasks retain their ID question families in the OOD split. Figure~\ref{fig:benchmark} shows representative ID and OOD examples.

\subsection{Data Generation and Evaluation}

\paragraph{Data generation and validation.} Each generator samples and renders a structured scene, then constructs a question and computes its answer from the scene. Task-specific checks assess whether the image contains the information needed to answer the question and whether the answer agrees with the scene. Scene structure and spatial annotations support data generation and validation. AdaLoop training uses only images, questions, and final answers.

\paragraph{Scoring.}
Scoring follows each task's answer format, without an LLM judge.
Labels, counts, and sequences use exact match after task-specific
parsing and normalization.
Ordered outputs must match the complete reference sequence,
while pair and index sets are compared independently of their order.
Tangram answers must match the unique reference placement, with each
rotation written as the smallest equivalent number of quarter turns.
We report per-task accuracy and macro averages across evaluated
tasks, separately for ID and OOD.

\subsection{Zero-Shot Results on All 24 Tasks}
\label{app:visro24-zero-shot}

\begin{table}[!t]
\centering
\caption{\textbf{Zero-shot results of API models on all 24 VisRO tasks.}
Protocol and Core6 responses as in the gray rows of Table~\ref{tab:main-results}; \emph{Overall}: macro average over tasks; \emph{tok.} as in Table~\ref{tab:main-results}.}
\label{tab:visro24-zero-shot}
\begingroup
\fontsize{8}{10}\selectfont
\setlength{\tabcolsep}{3.2pt}
\renewcommand{\arraystretch}{1.05}
\newlength{\visrotfnum}
\settowidth{\visrotfnum}{100.0}
\newlength{\visrotfa}\settowidth{\visrotfa}{Gemini 4 Argon}\setlength{\visrotfa}{\dimexpr\visrotfa/2-\tabcolsep\relax}\ifdim\visrotfa<\visrotfnum\setlength{\visrotfa}{\visrotfnum}\fi
\newlength{\visrotfb}\settowidth{\visrotfb}{GPT-6 Astra}\setlength{\visrotfb}{\dimexpr\visrotfb/2-\tabcolsep\relax}\ifdim\visrotfb<\visrotfnum\setlength{\visrotfb}{\visrotfnum}\fi
\newlength{\visrotfc}\settowidth{\visrotfc}{Claude Opus 5.5}\setlength{\visrotfc}{\dimexpr\visrotfc/2-\tabcolsep\relax}\ifdim\visrotfc<\visrotfnum\setlength{\visrotfc}{\visrotfnum}\fi
\newlength{\visrotfd}\settowidth{\visrotfd}{Grok 4.6}\setlength{\visrotfd}{\dimexpr\visrotfd/2-\tabcolsep\relax}\ifdim\visrotfd<\visrotfnum\setlength{\visrotfd}{\visrotfnum}\fi
\newlength{\visrotfe}\settowidth{\visrotfe}{DeepSeek-V4.1-Flash}\setlength{\visrotfe}{\dimexpr\visrotfe/2-\tabcolsep\relax}\ifdim\visrotfe<\visrotfnum\setlength{\visrotfe}{\visrotfnum}\fi
\begin{tabular*}{\linewidth}{@{\extracolsep{\fill}}l*{2}{>{\centering\arraybackslash}p{\visrotfa}}*{2}{>{\centering\arraybackslash}p{\visrotfb}}*{2}{>{\centering\arraybackslash}p{\visrotfc}}*{2}{>{\centering\arraybackslash}p{\visrotfd}}*{2}{>{\centering\arraybackslash}p{\visrotfe}}@{}}
\toprule
\multirow{2}{*}{Task} & \multicolumn{2}{c}{Gemini 4 Argon} & \multicolumn{2}{c}{GPT-6 Astra} & \multicolumn{2}{c}{Claude Opus 5.5} & \multicolumn{2}{c}{Grok 4.6} & \multicolumn{2}{c}{DeepSeek-V4.1-Flash} \\
\cmidrule(l{2pt}r{2pt}){2-3}\cmidrule(l{2pt}r{2pt}){4-5}\cmidrule(l{2pt}r{2pt}){6-7}\cmidrule(l{2pt}r{2pt}){8-9}\cmidrule(l{2pt}r{2pt}){10-11}
 & ID & OOD & ID & OOD & ID & OOD & ID & OOD & ID & OOD \\
\midrule
\multicolumn{11}{@{}l}{\emph{Discrimination and grouping}} \\
Count clusters & 98.3 & 88.3 & 88.3 & 73.3 & 86.7 & 63.3 & 36.7 & 6.7 & 56.7 & 13.3 \\
Points on lines & 100.0 & 95.0 & 100.0 & 96.7 & 100.0 & 88.3 & 0.0 & 0.0 & 0.0 & 0.0 \\
Find the same & 96.7 & 86.7 & 88.3 & 78.3 & 80.0 & 53.3 & 23.3 & 0.0 & 41.7 & 0.0 \\
Find the different & 100.0 & 100.0 & 98.3 & 88.3 & 98.3 & 100.0 & 61.7 & 41.7 & 76.7 & 58.3 \\
Find the shadow & 100.0 & 100.0 & 100.0 & 100.0 & 96.7 & 95.0 & 18.3 & 11.7 & 8.3 & 1.7 \\
Overlapping glyphs & 81.7 & 53.3 & 55.0 & 33.3 & 36.7 & 16.7 & 36.7 & 11.7 & 23.3 & 3.3 \\
Overlap matching & 48.3 & 25.0 & 93.3 & 53.3 & 31.7 & 15.0 & 21.7 & 16.7 & 28.3 & 16.7 \\
\midrule
\multicolumn{11}{@{}l}{\emph{Paths and connectivity}} \\
Flow Free & 95.0 & 90.0 & 95.0 & 100.0 & 55.0 & 36.7 & 0.0 & 0.0 & 3.3 & 0.0 \\
Metro map & 95.0 & 93.3 & 91.7 & 76.7 & 83.3 & 63.3 & 33.3 & 11.7 & 20.0 & 20.0 \\
Maze & 63.3 & 35.0 & 78.3 & 61.7 & 50.0 & 21.7 & 31.7 & 28.3 & 21.7 & 25.0 \\
Path tracing & 83.3 & 88.3 & 85.0 & 83.3 & 65.0 & 38.3 & 36.7 & 33.3 & 30.0 & 18.3 \\
\midrule
\multicolumn{11}{@{}l}{\emph{Planar transformations and composition}} \\
Jigsaw completion & 100.0 & 65.0 & 95.0 & 71.7 & 51.7 & 0.0 & 0.0 & 0.0 & 1.7 & 0.0 \\
Reconstruction & 51.7 & 30.0 & 90.0 & 58.3 & 60.0 & 28.3 & 33.3 & 18.3 & 35.0 & 21.7 \\
Tangram reconstruction & 100.0 & 98.3 & 98.3 & 90.0 & 96.7 & 36.7 & 13.3 & 0.0 & 20.0 & 0.0 \\
Mirroring patterns & 100.0 & 95.0 & 98.3 & 93.3 & 100.0 & 100.0 & 63.3 & 0.0 & 81.7 & 3.3 \\
Rotation patterns & 81.7 & 68.3 & 100.0 & 100.0 & 98.3 & 98.3 & 1.7 & 0.0 & 3.3 & 0.0 \\
Logic matrix & 98.3 & 98.3 & 96.7 & 20.0 & 71.7 & 0.0 & 28.3 & 0.0 & 28.3 & 0.0 \\
\midrule
\multicolumn{11}{@{}l}{\emph{Spatial transformations and structure}} \\
3D block counting & 65.0 & 8.3 & 93.3 & 88.3 & 25.0 & 6.7 & 5.0 & 0.0 & 8.3 & 1.7 \\
Block changes & 90.0 & 38.3 & 91.7 & 96.7 & 41.7 & 0.0 & 13.3 & 1.7 & 21.7 & 0.0 \\
Cube unfolding & 81.7 & 41.7 & 98.3 & 81.7 & 65.0 & 5.0 & 20.0 & 0.0 & 40.0 & 0.0 \\
Paper folding & 100.0 & 100.0 & 100.0 & 100.0 & 98.3 & 90.0 & 96.7 & 66.7 & 70.0 & 25.0 \\
3D views & 93.3 & 88.3 & 100.0 & 100.0 & 95.0 & 96.7 & 45.0 & 13.3 & 40.0 & 16.7 \\
View reconstruction & 96.7 & 95.0 & 98.3 & 98.3 & 76.7 & 81.7 & 28.3 & 38.3 & 40.0 & 13.3 \\
Viewpoint rotation & 83.3 & 70.0 & 90.0 & 91.7 & 48.3 & 33.3 & 30.0 & 35.0 & 23.3 & 25.0 \\
\midrule
\textbf{Overall} & 87.6 & 73.0 & \textbf{92.6} & \textbf{80.6} & 71.3 & 48.7 & 28.3 & 14.0 & 30.1 & 11.0 \\
tok. & \multicolumn{2}{c}{10{,}278} & \multicolumn{2}{c}{4{,}396} & \multicolumn{2}{c}{31{,}889} & \multicolumn{2}{c}{31{,}253} & \multicolumn{2}{c}{24{,}047} \\
\bottomrule
\end{tabular*}
\endgroup
\end{table}

Table~\ref{tab:visro24-zero-shot} extends the zero-shot evaluation
in Table~\ref{tab:main-results} to all 24 VisRO tasks.
GPT-6 Astra is the most accurate
($92.6\%$ ID, $80.6\%$ OOD), followed by Gemini~4~Argon
($87.6\%$, $73.0\%$) and Claude Opus~5.5 ($71.3\%$, $48.7\%$).
Grok~4.6 and DeepSeek-V4.1-Flash stay near $30\%$ ID and below
$15\%$ OOD.
Some tasks remain difficult out of distribution even for the strongest
models.

\begingroup
\newcommand{\benchpanel}[3]{%
  \begin{minipage}[t]{0.318\linewidth}
  \centering
  \parbox[b][\linewidth][c]{\linewidth}{\centering
    #1}\\[-1pt]
  {\small\textbf{#2}\par}
  \vspace{2pt}
  {\small #3\par}
  \end{minipage}}
\begin{figure}[p]
\centering
\benchpanel{\includegraphics[width=\linewidth,height=\linewidth,keepaspectratio]{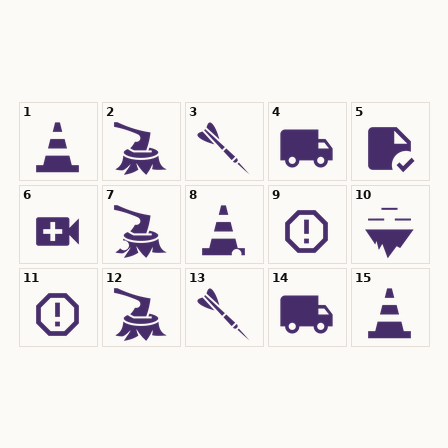}}{Find the same}{Find every identical pair of icons.}\hfill
\benchpanel{\includegraphics[width=\linewidth,height=\linewidth,keepaspectratio]{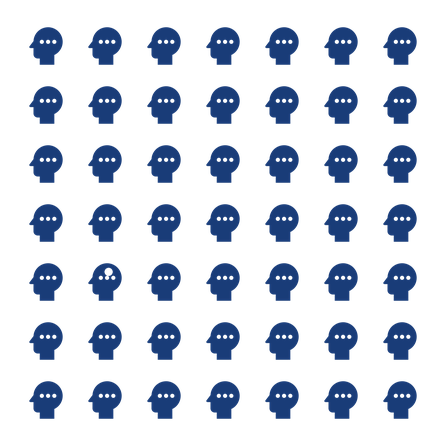}}{Find the different}{Which icon is different?}\hfill
\benchpanel{\includegraphics[width=\linewidth,height=\linewidth,keepaspectratio]{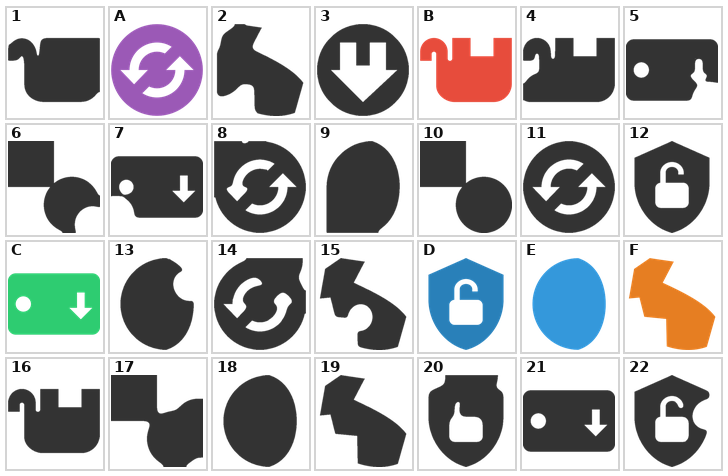}}{Find the shadow}{Match shadows; list the rest.}

\vspace{10pt}
\benchpanel{\includegraphics[width=\linewidth,height=\linewidth,keepaspectratio]{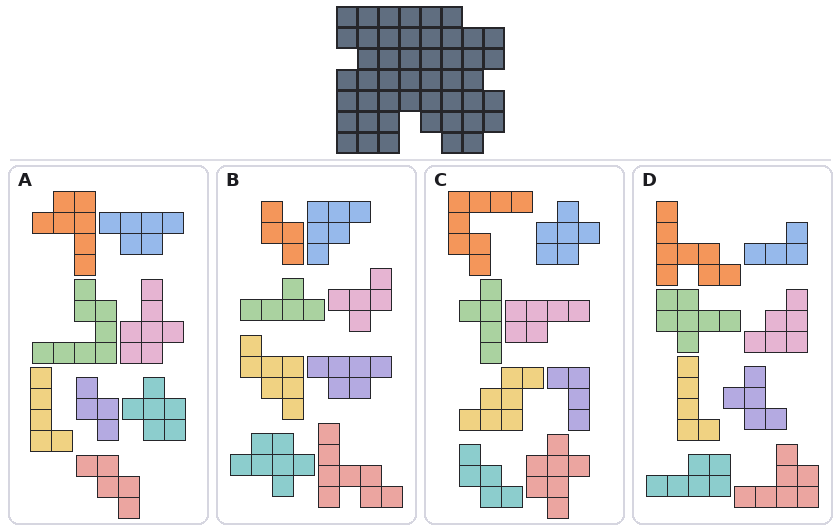}}{Reconstruction}{Which pieces tile the top shape?}\hfill
\benchpanel{\includegraphics[width=\linewidth,height=\linewidth,keepaspectratio]{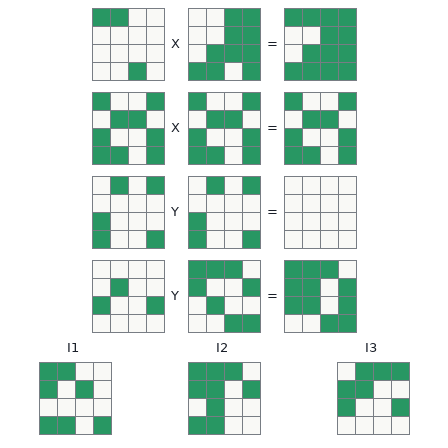}}{Logic matrix}{Infer X, Y; give the final grid.}\hfill
\benchpanel{\includegraphics[width=\linewidth,height=\linewidth,keepaspectratio]{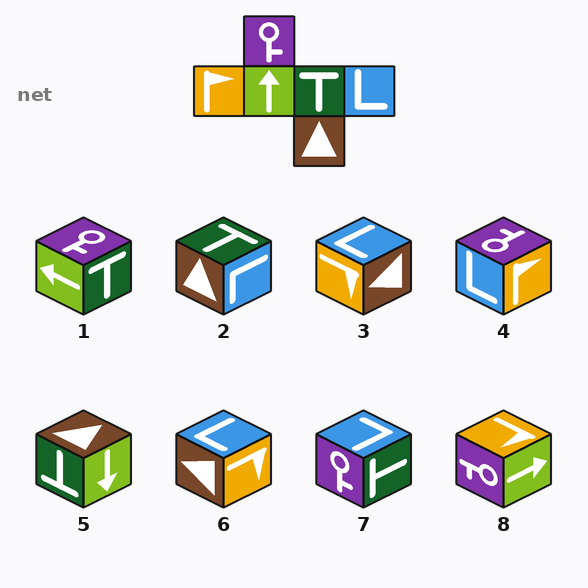}}{Cube unfolding}{Which cubes match the folded net?}

\vspace{10pt}
\benchpanel{\includegraphics[width=\linewidth,height=\linewidth,keepaspectratio]{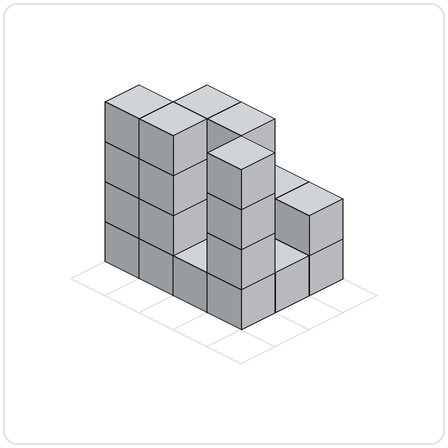}}{3D block counting}{Find the minimum cube count.}\hfill
\benchpanel{\includegraphics[width=\linewidth,height=\linewidth,keepaspectratio]{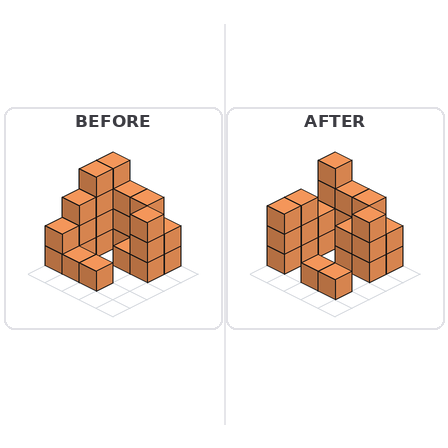}}{Block changes}{Count added and removed cubes.}\hfill
\benchpanel{\includegraphics[width=\linewidth,height=\linewidth,keepaspectratio]{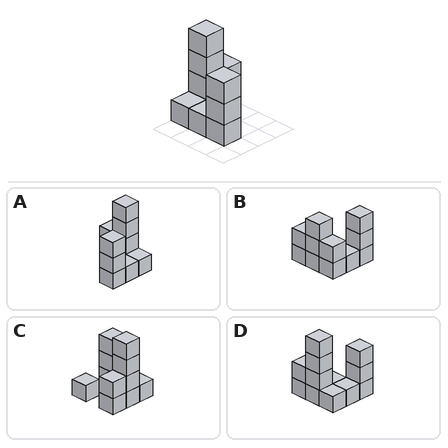}}{Viewpoint rotation}{Find the same object after rotation.}
\caption{\textbf{VisRO task examples (1/2).}
The examples illustrate matching and discrimination (top), composition and
transformations (middle), and spatial structure (bottom). All examples come
from the \emph{hard} level of the released ID evaluation split;
the text below each image summarizes its question.}
\label{fig:benchmark-extended}
\end{figure}

\begin{figure}[p]
\centering
\benchpanel{\includegraphics[width=\linewidth,height=\linewidth,keepaspectratio]{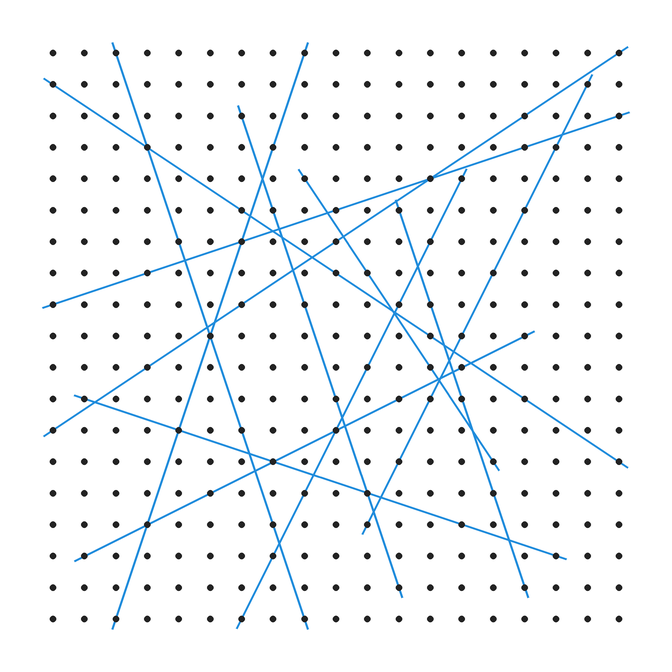}}{Points on lines}{Count per-line and distinct dots.}\hfill
\benchpanel{\includegraphics[width=\linewidth,height=\linewidth,keepaspectratio]{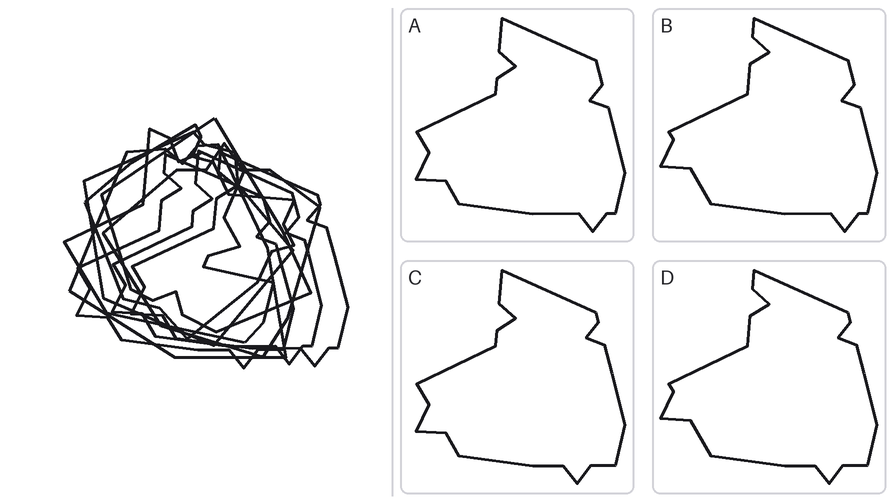}}{Overlap matching}{Which shape is in the overlap?}\hfill
\benchpanel{\includegraphics[width=\linewidth,height=\linewidth,keepaspectratio]{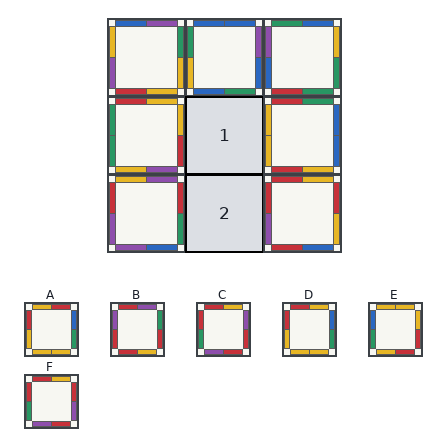}}{Jigsaw completion}{Fill the holes with turned pieces.}

\vspace{10pt}
\benchpanel{\includegraphics[width=\linewidth,height=\linewidth,keepaspectratio]{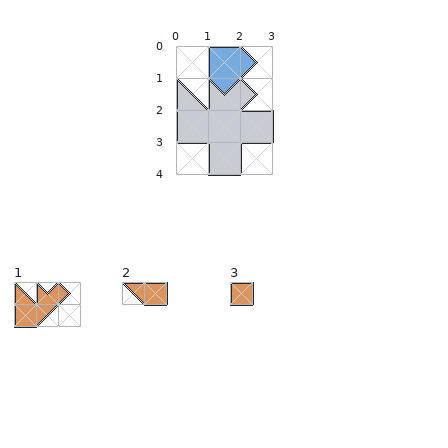}}{Tangram reconstruction}{Place the pieces to cover the target.}\hfill
\benchpanel{\includegraphics[width=\linewidth,height=\linewidth,keepaspectratio]{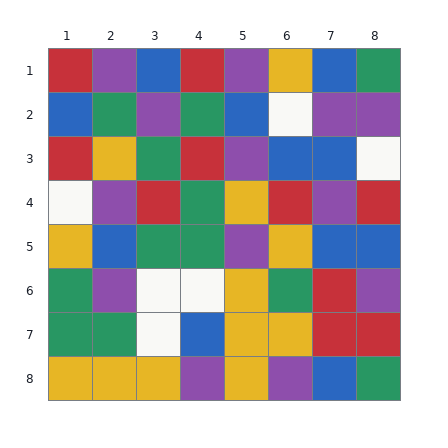}}{Mirroring patterns}{Apply reversals; read four cells.}\hfill
\benchpanel{\includegraphics[width=\linewidth,height=\linewidth,keepaspectratio]{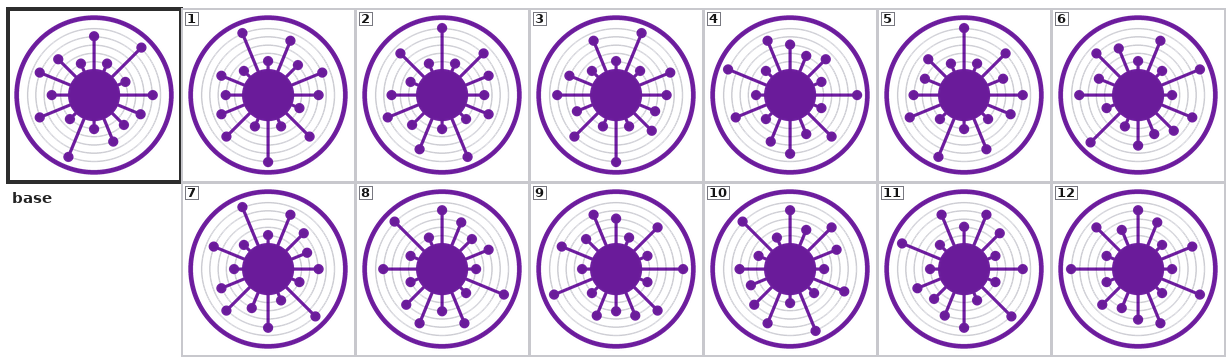}}{Rotation patterns}{Which figure matches each angle?}

\vspace{10pt}
\benchpanel{\includegraphics[width=\linewidth,height=\linewidth,keepaspectratio]{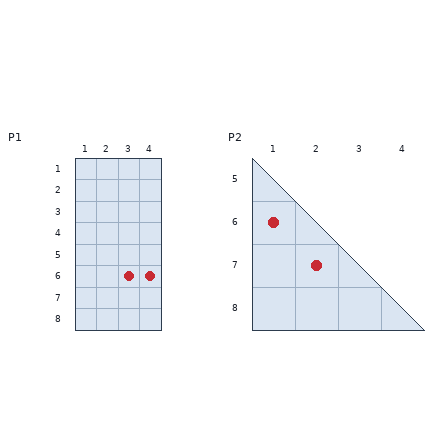}}{Paper folding}{Fold, punch, unfold; locate holes.}\hfill
\benchpanel{\includegraphics[width=\linewidth,height=\linewidth,keepaspectratio]{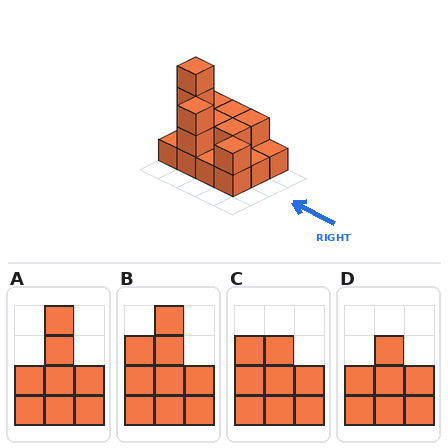}}{3D views}{Find the right-side view.}\hfill
\benchpanel{\includegraphics[width=\linewidth,height=\linewidth,keepaspectratio]{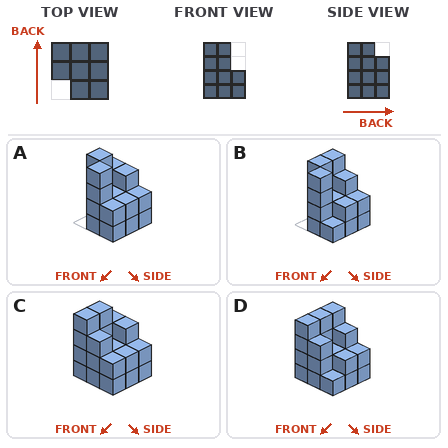}}{View reconstruction}{Which structure fits the views?}
\caption{\textbf{VisRO task examples (2/2).}
The examples illustrate point counting, overlap matching, and
jigsaw completion (top), planar composition and transformations (middle),
and folding and three-dimensional views (bottom). As in
Figure~\ref{fig:benchmark-extended}, all examples come from the \emph{hard}
level of the released ID evaluation split, and the text summarizes each question.}
\label{fig:benchmark-extended-more}
\end{figure}
\endgroup

\end{document}